\documentclass[lettersize,journal]{IEEEtran}
\usepackage{cite}
\usepackage{amsmath,amssymb,amsfonts}
\usepackage{multicol,listings,multirow}
\usepackage{algorithmic}
\usepackage{graphicx}
\usepackage{textcomp}
\usepackage{array}
\usepackage[colorlinks=true,
            linkcolor=blue,
            filecolor=blue,      
            urlcolor=blue,       
            citecolor=blue
            ]{hyperref}
\usepackage[marginal]{footmisc}
\usepackage[misc]{ifsym}
\usepackage{booktabs}
\usepackage{color}
\usepackage[table]{xcolor}
\usepackage{overpic}
\usepackage[ruled,linesnumbered]{algorithm2e}
\usepackage{flushend}
\usepackage{makecell}

\def\BibTeX{{\rm B\kern-.05em{\sc i\kern-.025em b}\kern-.08em
    T\kern-.1667em\lower.7ex\hbox{E}\kern-.125emX}}
\begin{document}
\vspace{-0.5cm}
\title{FedLPPA: Learning Personalized Prompt and Aggregation for Federated Weakly-supervised Medical Image Segmentation}
\author{Li Lin, Yixiang Liu, Jiewei Wu, Pujin Cheng, Zhiyuan Cai, Kenneth K. Y. Wong, \IEEEmembership{Senior Member, IEEE}, Xiaoying Tang, \IEEEmembership{Senior Member, IEEE} 
\vspace{-0.8cm}
\thanks{Li Lin, Yixiang Liu and Jiewei Wu contributed equally to this work.}
\thanks{Li Lin, Yixiang Liu, Jiewei Wu and Xiaoying Tang are with the Department of Electronic and Electrical Engineering, Southern University of Science and Technology (SUSTech), Shenzhen, China and Jiaxing Research Institute, Southern University of Science and Technology, Jiaxing, China. Li Lin, Pujin Cheng and Kenneth K. Y. Wong are with the Department of Electrical and Electronic Engineering, the University of Hong Kong, Hong Kong, China (linli@eee.hku.hk). Zhiyuan Cai is with the Department of Computer Science and Engineering, the Hong Kong University of Science and Technology, Hong Kong, China. Corresponding author: Xiaoying Tang (tangxy@sustech.edu.cn).}
\thanks{This study was supported by the National Key Research and Development Program of China (2023YFC2415400); National Natural Science Foundation of China (62071210); Shenzhen Science and Technology Program (RCYX20210609103056042); Shenzhen Science and Technology Innovation Committee (KCXFZ2020122117340001); Guangdong Basic and Applied Basic Research (2021A1515220131).
}
}

\maketitle

\widowpenalty=5
\clubpenalty=5
\setlength\abovedisplayskip{0.13cm}
\setlength\belowdisplayskip{0.13cm}

\vspace{-0.5cm}

\begin{abstract}

Federated learning (FL) effectively mitigates the data silo challenge brought about by policies and privacy concerns, implicitly harnessing more data for deep model training. 
However, traditional centralized FL models grapple with diverse multi-center data, especially in the face of significant data heterogeneity, notably in medical contexts.
In the realm of medical image segmentation, the growing imperative to curtail annotation costs has amplified the importance of weakly-supervised techniques which utilize sparse annotations such as points, scribbles, etc.
A pragmatic FL paradigm shall accommodate diverse annotation formats across different sites, which research topic remains under-investigated.
In such context, we propose a novel personalized FL framework with learnable prompt and aggregation (FedLPPA) to uniformly leverage heterogeneous weak supervision for medical image segmentation. In FedLPPA, a learnable universal knowledge prompt is maintained, complemented by multiple learnable personalized data distribution prompts and prompts representing the supervision sparsity. Integrated with sample features through a dual-attention mechanism, those prompts empower each local task decoder to adeptly adjust to both the local distribution and the supervision form. Concurrently, a dual-decoder strategy, predicated on prompt similarity, is introduced for enhancing the generation of pseudo-labels in weakly-supervised learning, alleviating overfitting and noise accumulation inherent to local data, while an adaptable aggregation method is employed to customize the task decoder on a parameter-wise basis. Extensive experiments on {four} distinct medical image segmentation tasks involving different modalities underscore the superiority of FedLPPA, with its efficacy closely parallels that of fully supervised centralized training.
Our code and data will be available at \hyperlink{code}{https://github.com/llmir/FedLPPA}.


\end{abstract}

\begin{IEEEkeywords}
Federated learning, Prompt-driven personalization, Heterogeneous weak supervision, Learnable aggregation, Medical image segmentation
\end{IEEEkeywords}

\section{Introduction}
\label{sec:introduction}

%
\IEEEPARstart{M}{edical} image segmentation is a pivotal component in computer-aided diagnosis, treatment planning, and follow-ups. It aims to delineate regions of interest within medical images, such as physiological structures or pathological regions, facilitating subsequent analyses \cite{wang2022medical}. In recent years, deep learning (DL) techniques have dominated this task due to their superior performance. A multitude of studies have been introduced, primarily focusing on devising sophisticated network architectures or formulating mathematically rigorous and topology-preserving loss functions \cite{shit2021cldice,lin2021bsda}. DL methods largely benefit from large-scale fully annotated datasets which nevertheless are becoming increasingly infeasible in practical clinical contexts for two primary reasons: 1) Due to privacy and security concerns, numerous regulations and guidelines mandate the protection of patient data, limiting sharing across medical institutions (the data silo issue) \cite{goddard2017eu}; 2) Densely/Fully annotating a vast array of medical data is both prohibitively expensive and time-consuming since it requires specialized expertise. As the array of imaging modalities and clinical tasks proliferates, adopting sparse annotations or labeling only a fraction of the data emerges as more pragmatic approaches.

\begin{figure}[t]
    \centerline{\includegraphics[width=0.96\columnwidth]{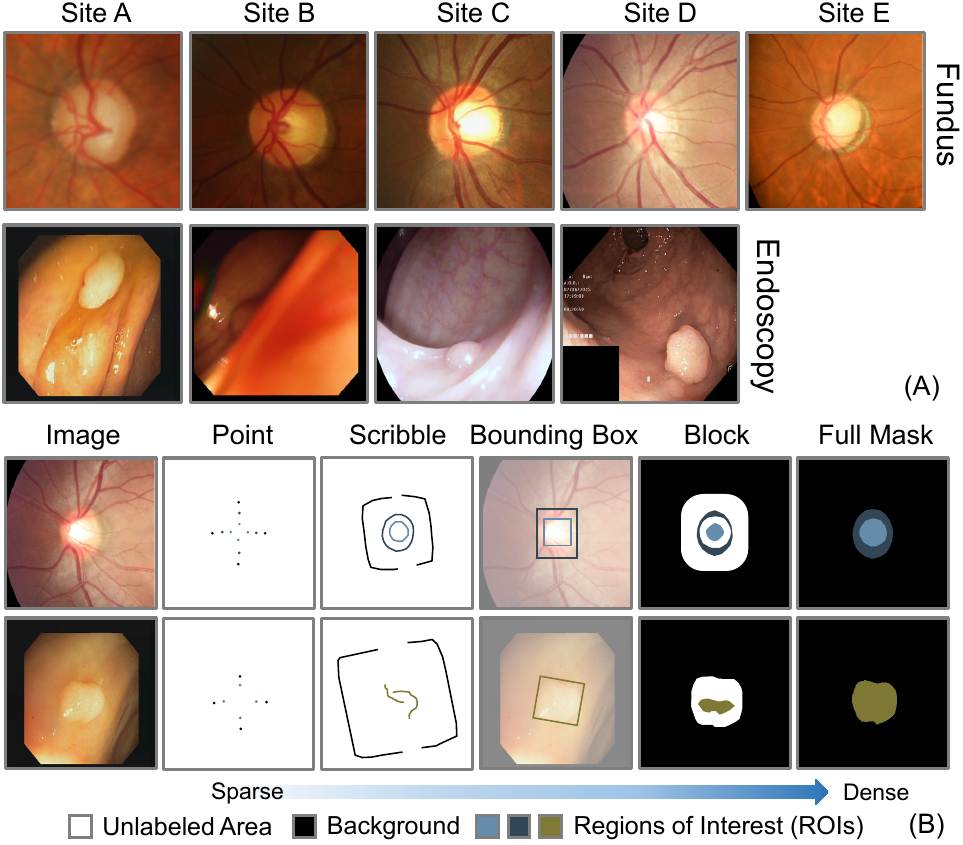}}
    \vspace{-0.1cm}
    \caption{A: Data samples from different centers showcase the domain gaps in their distributions; B: Examples of typical weak labels and their corresponding masks, with annotation granularity from sparse to dense.}
    \label{fig1}
    \vspace{-0.5cm}
\end{figure}

To address the data silo issue, federated learning (FL) has emerged and garnered extensive attention, as it allows different centers to collaboratively train powerful deep models without the need to share or centralize their data \cite{li2020federated}.
In standard FL paradigms, each client uses its local data to train a model and collaborates by aggregating individual local model parameters to a central server in some manner of aggregation and broadcasting the updated parameters to each individual site. For instance, the prevailing FedAvg averages models from all sites according to sample weights \cite{mcmahan2017communication}. However, when confronted with the significant challenge of statistical heterogeneity in federated settings, despite some efforts in trying to augment the generalization of the global model through employing style augmentation or enhancing aggregation strategies \cite{liu2021feddg,li2020federated2,wicaksana2022fedmix}, studies have shown that utilizing a solitary global model rarely yields optimal results across all institutions. Such {\textbf{data heterogeneity}} is further exacerbated in medical image segmentation tasks, wherein differences in imaging equipments, protocols, patient populations, and physician expertises lead to even more diverse domain biases, as shown in Fig. \ref{fig1}A.

In this context, personalized federated learning (pFL) proposes to learn personalized local models for individual sites to accommodate unique feature distributions or customized prediction patterns \cite{tan2022towards}. Existing pFL paradigms mainly encompass approaches through local fine-tuning \cite{wang2019federated}, segmenting the network and retaining personalization layer parameters for local models \cite{collins2021exploiting,li2020fedbn}, similarity-based aggregation \cite{lu2022personalized}, or knowledge distillation \cite{chen2023metafed}. However, these approaches might suffer from catastrophic forgetting of public knowledge and overfitting to local distributions, inability to leverage valuable similarities from other sites, or unsuitable personalization due to coarse-grained aggregation \cite{lin2023unifying,zhang2023fedala}. Recently, prompt-based approaches have gained prominence in the fields of natural language processing and computer vision \cite{liu2023pre,jia2022visual,liu2023clip}, where a set of vectors, either manually set or learnt, are used as conditions for a model to better perceive contextual information such as different data distributions or downstream tasks. This allows for uniform training on heterogeneous data (different distributions, labelling forms, tasks, etc.) with only modest increases in computational costs, but is still under-researched in federated medical image segmentation. 

Concurrently, the pressing need to reduce annotation costs has heightened the focus on weakly supervised segmentation (WSS). WSS techniques typically leverage sparse granularity supervision (like points, boxes, scribbles, or blocks, as shown in Fig. \ref{fig1}B) through novel loss function designing, consistency learning, adversarial learning, or data synthesis \cite{liang2022tree,lin2023yolocurvseg,zhang2022cyclemix,valvano2021learning}. Integrating WSS into FL can further reduce the annotation cost per site while leveraging the benefits of FL. However, there is little research on WSS in pFL. 
A more challenging but practical setup is to allow for different sites to possess different forms of weak labels, which further introduces {\textbf{label/supervision heterogeneity}}. 
{In this setting, the challenges include data heterogeneity, label heterogeneity, and insufficient supervision signals. These issues collectively lead to uncoordinated model training rates or progress across multiple centers in FL, complicating interaction and coordination.}
In such context, it is highly desirable to design a unified and easy-to-deploy pFL framework for WSS settings.

In this paper, we extend our previous work \cite{lin2023unifying} and propose a novel pFL framework with learnable prompt and aggregation (FedLPPA) to uniformly leverage heterogeneous weak supervision for medical image segmentation. 
In FedLPPA, prompts representing universal knowledge, data distribution, and annotation sparsity are maintained, with the former two being learnable whereas the latter is manually set. These personalized prompts are fused with data representations and perceived by decoders through a dual-attention (spatial and channel) fusion mechanism. A personalized dual-decoder mechanism based on prompts' similarity, coupled with learnable aggregation, is introduced to efficiently and adaptively generate pseudo-labels for each site, thereby facilitating WSS.
Our main contributions are summarized as follows:
\begin{itemize}
    \item We tackle a practical but challenging federated heterogeneous WSS problem via our FedLPPA framework, which to our knowledge is a pioneering work in its own setting.
    \item A Tri-prompt Dual-attention Fusion (TDF) module is proposed for unified and lightweight in-context learning, enabling each local model to adapt to unique data distribution and supervision sparsity for personalized learning. 
    \item A Prompt similarity Dual-decoder with Learnable Aggregation (PDLA) mechanism is designed to obtain fine personalized parameters, alleviating local noise accumulation and generating superior pseudo-proposals for WSS.
    \item Our FedLPPA establishes consistent superiority over general FL and state-of-the-art pFL methods in extensive experiments on {four} different medical image segmentation tasks involving different modalities.
\end{itemize}

{This paper extends our FedICRA \cite{lin2023unifying} in several aspects:
\begin{itemize}
    \item Discarding the single one-hot prompt, which fails to finely cope with diverse training conditions and misses cross-client insights, we introduce a design using triple latent prompts, two of which are learnable. This allows the model and prompts to more finely perceive and coordinate various training conditions/contexts, while maintaining and representing inter-client common knowledge and relationship. Additionally, we perform multiple fusion of prompts and image features to better enhance the features.
    \item Forgoing the multi-level tree affinity WSS paradigm, we propose a PDLA paradigm, which reduces computational overhead. By leveraging prompts to personalize the selection or aggregation of auxiliary decoders, this approach effectively gathers useful information from similar clients and reduces interference. This strategy assists in generating higher-quality and more diverse pseudo-labels, thereby enhancing the segmentation performance.
    \item We validate the performance of FedLPPA in a broader range of medical image segmentation scenarios, including endoscopy and MRI. Additionally, we conduct more ablation and analysis experiments to further scrutinize the efficacy of the proposed framework.
\end{itemize}
}

\section{Related Work}


\subsection{Federated Learning for Medical Image Analysis}
\vspace{-0.05cm}

FL is a multi-center training paradigm developed to tackle the data silo issue by training a robust global model (general FL, gFL) or multiple personalized models for individual sites (pFL) without data sharing nor centralization \cite{li2020federated}. In the wake of FedAvg, most gFL methods in medical image analysis mitigate the negative effects of heterogeneity through style augmentation, aggregation strategy enhancement, and disentangled representation, thereby increasing the global model's generalizability \cite{mcmahan2017communication,liu2021feddg,li2020federated2,wicaksana2022fedmix,yan2020variation,bercea2022federated}. However, employing a single global model is generally suboptimal; thus, pFL emerges as a more promising and flexible paradigm, customizing local models to better accommodate the unique distributions and needs of individual sites\cite{tan2022towards}. Existing approaches achieve this objective through local fine-tuning, model decoupling, knowledge distillation, similarity aggregation or inconsistency exploitation \cite{wang2019federated,collins2021exploiting,li2020fedbn,wang2022personalizing}. However, these methods may suffer from catastrophic forgetting of common knowledge, inability to utilize information from other sites, inappropriate personalization, or high computational overhead. 

To reduce annotation costs, efforts have been made to investigate FL methods in semi-supervised, self-supervised and weakly supervised settings \cite{liu2021federated,wu2023federated,yan2023label,wu2022distributed,zhu2023feddm,pan2023pfedwsd}. Among them, Liu et al. \cite{liu2021federated} and Wu et al. \cite{wu2023federated} respectively facilitate the effective use of data and knowledge transfer between labeled and unlabeled sites through inter-client relation matching and prototype-based pseudo-labeling. However, this semi-suppervised FL setting is not conducive to fairness among federated sites, leading to potential free-riders (unlabeled clients). Moreover, semi-supervised learning is notably influenced by the gap between labeled and unlabeled data, not to mention the more pronounced domain shifts across different centers. Contrastive learning and masked image modelling techniques are applied to self-supervised settings under FL, yet they still demand fine-tuning under certain amounts of labeled data, and the performance gain compared to fully supervised counterparts remains slight \cite{yan2023label,wu2022distributed}. Meanwhile, Zhu et al. \cite{zhu2023feddm} and Pan et al. \cite{pan2023pfedwsd} are pioneers in integrating WSS and FL.
Zhu et al. \cite{zhu2023feddm}, belonging to the general category of gFL, focus on bounding box labels which are hard to align with other types of weak labels (see Sec. \ref{subsec:wss}). Pan et al. \cite{pan2023pfedwsd} do not delve into leveraging various weak label forms. 
In contrast, our work focuses on heterogeneous WSS in pFL and employs a learnable prompt-based in-context learning approach to differentiate and unify various annotations.

\vspace{-0.1cm}
\subsection{Singular/Hybrid Weakly-supervised Segmentation}
\label{subsec:wss}
WSS is a paradigm that trains dense segmentation models using sparse labels. Current work in WSS primarily focuses on designs for specific types of weak annotations, commonly involving image-, point-, scribble-, bounding box-, and block-wise supervision \cite{liang2022tree,lin2023yolocurvseg,zhang2022cyclemix,valvano2021learning,kervadec2020bounding,zhang2023weakly,chen2022c}. Within this scope, image-level supervised approaches, typically reliant on classification network features like Class Activation Maps (CAMs) \cite{chen2022c}, are not suitable for segmenting organs and structures that are consistently presented in images, leading to exclusion of image-level annotations from our consideration. On the other hand, unlike point, scribble, or block annotations, bounding boxes supervise position and range rather than pixel categories. Methods developed for box labels typically involve constraints, priors, and network architectures that differ substantially from those for other types of weak labels \cite{kervadec2020bounding,wang2021bounding}, making it difficult to be accommodated nor heterogeneously utilized in a unified manner across different sites in a federated setting \cite{zhu2023feddm}. The framework we propose accommodates all weak supervision forms except the image-level one. And we implement specifically-designed preprocessing steps before model training to convert boxes into other types of weak labels. 

Generally, apart from those for image-level labels, WSS methods predominantly involve paradigms based on pseudo-proposal, consistency learning, adversarial learning, and data synthesis, etc \cite{liang2022tree,lin2023yolocurvseg,zhang2022cyclemix,valvano2021learning}. Among these, pseudo-proposal methods stand out for their universality and flexibility, utilizing priors, semantic similarities, or model predictions to extend sparse labels to unlabeled regions and generate pseudo-labels \cite{liang2022tree}. However, since weak labels typically lack direct supervision at boundary areas, models are prone to being constrained by their own predictions, especially in uncertain regions, which can lead to update stagnation or noise accumulation \cite{zhang2021adaptive,luo2022scribble}. Our method employs a prompt-similarity based dual-decoder mechanism to alleviate this issue by leveraging useful information from similar sites.

To maximize compatibility across various annotation types and utilize more data, hybrid supervision frameworks have also garnered considerable research interest. Luo et al. \cite{luo2021oxnet} and Chai et al. \cite{chai2022orf} each introduce a hybrid supervision framework for detecting thoracic diseases from X-rays and identifying rib fractures from CT scans. The former integrates weakly-labeled data in bounding box (Bbox) and image-level forms as well as unlabeled one, while the latter includes weakly-labeled data in Bbox and point forms as well as unlabeled one. Both are crafted for centralized settings and feature distinct network architectures and loss function designs for different supervision forms. Xu et al. \cite{xu2022anti} design a vessel segmentation framework based on the mean teacher paradigm, being compatible with both fully supervised and noisily supervised one. FedMix \cite{wicaksana2022fedmix} introduces an FL framework that is compatible with fully supervised, weakly supervised in Bbox and image-level forms, as well as unsupervised data, paralleling our research motivation. Yet, it necessitates different model and loss designs for different supervision types and overlooks domain shifts across different sites, falling within the gFL paradigm. In contrast, our goal is to develop a lightweight, unified pFL framework for all sites, accommodating all supervision forms except the less common image-level one.

\vspace{+0.1cm}
\section{Method} 


\begin{figure*}[t]
    \vspace{-0.2cm}
    \centering
    \includegraphics[width=\textwidth]{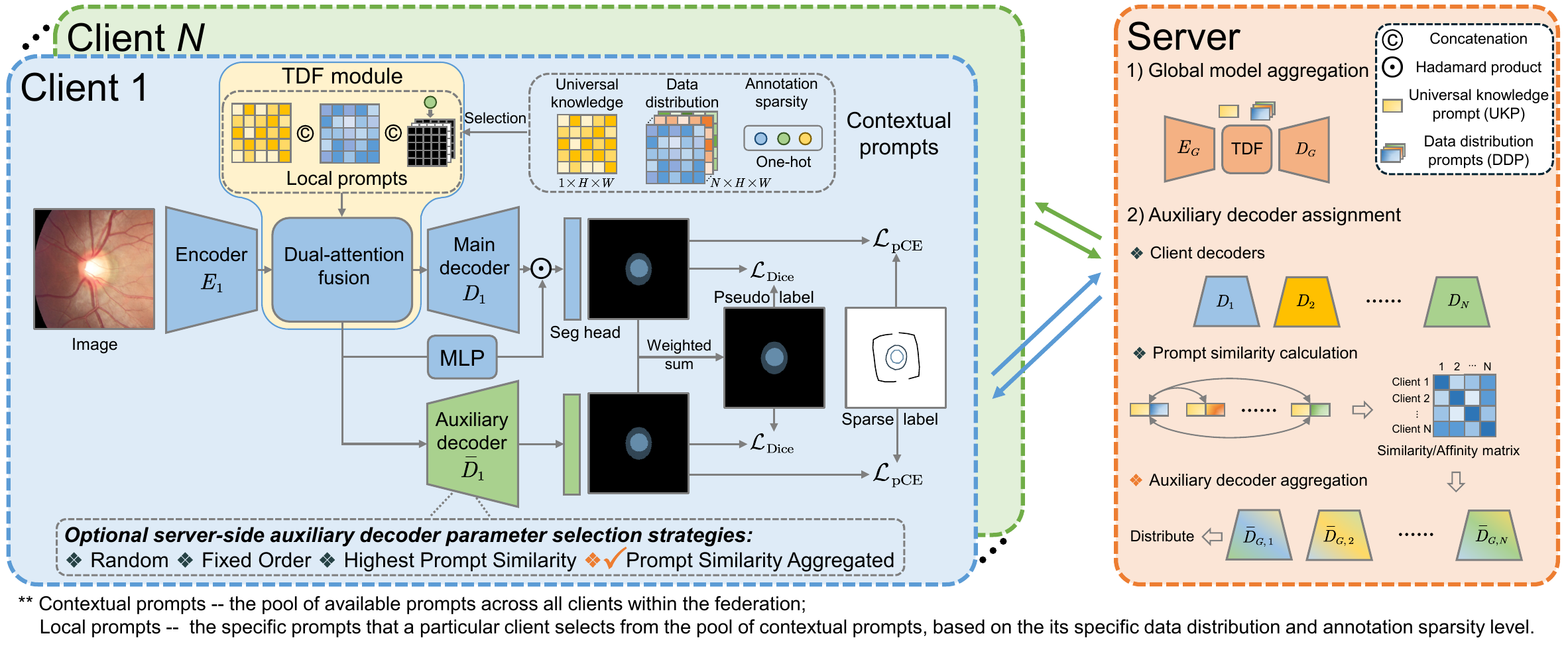}
    \vspace{-0.52cm}
    \caption{{Schematic illustration of the proposed FedLPPA framework. Note that the segmentation heads and the MLP block are incorporated within the decoders and are not depicted independently, except in the left panel of the scheme. The \textit{auxiliary decoder aggregation} on the right panel builds its basis on the selection of the \textit{Prompt Similarity Aggregated} strategy.
    }}
    \label{fig2}
    \vspace{-0.4cm}
\end{figure*}

\begin{figure}[!h]
    \centering
    \includegraphics[width=\columnwidth]{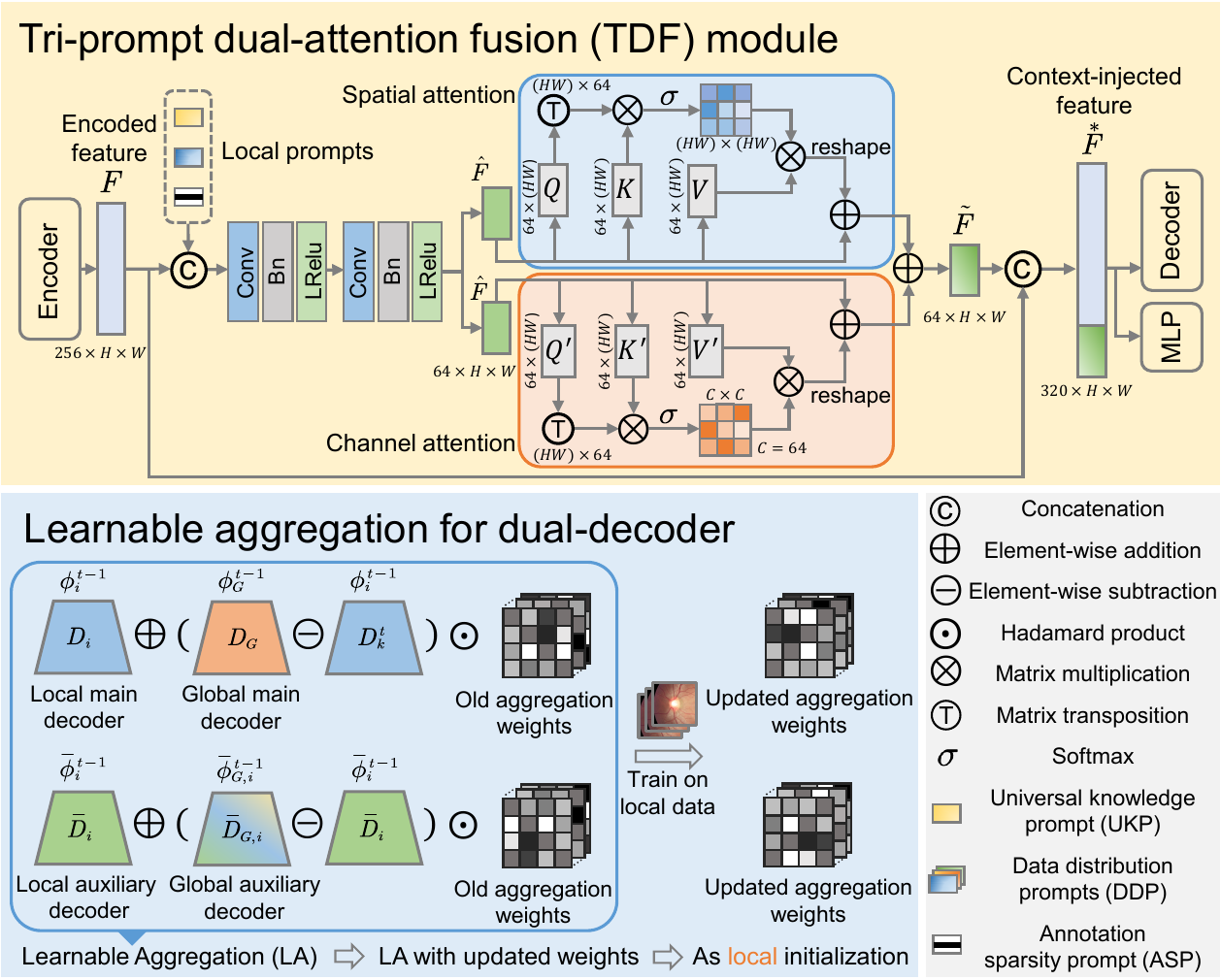}
    \vspace{-0.4cm}
    \caption{{Detailed illustration of the Tri-prompt Dual-attention Fusion (TDF) module and the Learnable Aggregation (LA) for dual-decoder mechanism. The term `context-injected' denotes that the training conditions or contextual information have been integrated into the original image features, thereby resulting in enhanced features.
    }}
    \label{new-fig3}
    \vspace{-0.3cm}
\end{figure}

In this section, we first briefly overview our FedLPPA framework, then detail the Tri-prompt Dual-attention Fusion (TDF) module and the Prompt similarity Dual-decoder with Learnable Aggregation (PDLA) mechanism, which are the two key components of FedLPPA.
The former leverages in-context learning to enable the task decoders and heads to perceive common knowledge across multiple centers, local data distribution, and supervision granularity for better adaptive adjustments. The latter personalizes the auxiliary decoder's parameters from sites with similar local distributions to mitigate noise accumulation and generate superior pseudo-labels, while learnable aggregation further personalizes the parameters of the dual-decoder. 
Additionally, we describe preprocessing operations for bounding box data to align with other supervision forms.
{The overall framework and key components of FedLPPA are respectively provided in Fig. \ref{fig2} and Fig. \ref{new-fig3}. }

\vspace{-0.1cm}
\subsection{FedLPPA Paradigm Overview}
With the support of a central server, FedLPPA is designed to establish a personalized model denoted as $\{{\theta}_i\}^N_{i=1}$ for each client/site without the necessity of data sharing, as opposed to constructing a single shared model in gFL methods. These models not only incorporate beneficial information from other sites but also adaptively fit to the corresponding local data distribution and supervision form. Assume there are $N$ sites with private data $\{\mathcal{D}_1,...,\mathcal{D}_N\}$ where $\mathcal{D}_i=\{(\mathbf{X}_i,\mathbf{Y}_i)\}_{i=1}^{N}$ with $\mathbf{X}_i$ and $\mathbf{Y}_i$ respectively denoting the data and labels of the $i^{th}$ site. The data distributions vary, i.e., $\mathcal{P}(\mathcal{D}_i)\neq\mathcal{P}(\mathcal{D}_j)$, and the annotation forms of $\{\mathbf{Y}_i\}_{i=1}^{N}$ may differ, including types such as point, scribble, bounding box and block. In FedLPPA, the global model and local models share the same network architecture, which is a U-shape network with an additional auxiliary decoder and a TDF module inserted in the middle. Furthermore, a multilayer perceptron (MLP) block links the TDF module with the main decoder's segmentation head (as shown in Fig. \ref{fig2}A), enabling more direct transfer of context-injected features and enhancing flexibility in parameter optimization. 
{We set the encoder and the TDF module (including the learnable prompts) as the globally shared parts, denoted by parameter $\vartheta$}. The main decoder (with MLP, parameter $\phi$) and the auxiliary decoder (parameter $\bar{\phi}$) are configured as personalized components. The training process of FedLPPA involves $T$ rounds of alternating updates between clients and server. In the $t^{th}$ round, all clients receive the same aggregated global parameters ($\vartheta_G^{t-1}$ and $\phi_G^{t-1}$) from the server, as well as customized auxiliary decoder parameter ($\bar{\phi}_{G,i}^{t-1}$) obtained according to the selected strategy. {The global part of the local model is initialized using $\vartheta_G^{t-1}$, while the personalized components are initialized with $\hat{\phi}_{i}^{t}$ and $\hat{\bar{\phi}}_{i}^{t}$, respectively. These parameters ($\hat{\phi}_{i}^{t}$ and $\hat{\bar{\phi}}_{i}^{t}$) are derived using the {Learnable Aggregation (LA)} mechanism based on the parameters distributed by the server and the local parameters from the previous round (i.e., (${\phi}_{G}^{t-1}$,${\phi}_{i}^{t-1}$) and ($\bar{\phi}_{G,i}^{t-1}$,$\bar{\phi}_{i}^{t-1}$), respectively).}
Each client updates its local model via optimizing the local objective with local data $\mathcal{D}_i$ and the manually set annotation sparsity prompt (ASP) $p_{S,i}$ used in TDF,
\begin{equation}
{\theta}_{i}^{t} = \arg\min_{{\theta}_{i}} \mathcal{L_\text{wss}}(\mathcal{D}_i, \vartheta_{G}^{t-1}, \hat{\phi}_{i}^{t}, \hat{\bar{\phi}}_{i}^{t}, p_{S,i}),
\end{equation}
where $\mathcal{L_\text{wss}}$ is an WSS loss function. After completing the local updates, all clients upload the current round's parameters ${\theta}_{i}^{t}$ to the server for global model aggregation and auxiliary decoder personalized parameter assignment,
\begin{equation}
\vartheta_{G}^{t}=\sum_{i=1}^N{\omega ^i}\vartheta_{i}^{t};\         \phi _{G}^{t}=\sum_{i=1}^N{\omega ^i}\phi _{i}^{t},
\end{equation}
where $\omega ^i=|\mathcal{D}_i|/\sum_{j=1}^N|\mathcal{D}_j|$ is the aggregation weight and $|\mathcal{D}_i|$ is the sample size of the $i^{th}$ site. The server-side auxiliary decoder parameters $\bar{\phi}_{G,i}^{t}$ will be introduced in Sec. \ref{subsec:PDLA}. {Algorithm \ref{algorithm} provides a detailed description of FedLPPA's training process.}

\begin{algorithm}[t]
    {\caption{FedLPPA}\label{algorithm}
    \KwIn{$\{\mathcal{D}_i\}_{i=1}^{N}$: training data of $N$ clients; $T$: total federation rounds; $p_{S,i}$: ASP}
    \KwOut{$\{\theta_i\}_{i=1}^{N}$: personalized models of $N$ clients}

    $\theta_G^0\leftarrow\mathbf{initialize()}$ \hfill $\triangleright \mathbf{Initialization}$\\
    Server sends $\theta_G^0$ to all clients, $\theta_i^0\leftarrow \theta_G^0$\\
    Client $i$ selects $p_{D,i}^0$, $p_{S,i}$ and initializes $W_{M,i}$, $W_{A,i}$ \\
    \If{$\mathcal{D}_i$ uses box label}{
        Perform predefined label preprocessing\\
    } 
    \For{$t=1:T$}{
        \While{$W_{M,i}$, $W_{A,i}$ do not converge}{
            Train $W_{M,i}$, $W_{A,i}$ with Eqs. (12-13) \hfill $\triangleright \mathbf{LA}$\\
            Clip $W_{M,i}$, $W_{A,i}$ to $[0,1]$\\
        }
        Obtain parameters $\hat{\phi}_i^t$, $\hat{\bar{\phi}}_i^t$ with Eqs. (10-11)\\
            Train $\theta_i^t$ with Eq. (1) \hfill $\triangleright \textbf{Local training}$\\
        Client $i$ uploads $\vartheta_{i}^{t}$, $\phi _{i}^{t}$ to server \hfill $\triangleright \textbf{Uploading}$\\
        Server obtains $\vartheta_G^t$, $\phi_G^t$ with Eq. (2)\\
        Calculate affinity matrix $A$ with Eqs. (7-8)\\
        Aggregate $\bar{\phi}_{G,i}^{t}$ with Eq. (9) (default strategy)\
        Distribute $\vartheta_G^t$, $\phi_G^t$ to all clients \hfill $\triangleright \textbf{Downloading}$\\
        Distribute $\{\bar{\phi}_{G,i}^{t}\}_{i=1}^{N}$ to corresponding clients\\
    }
    \Return $\{\theta_i\}_{i=1}^{N}$\\}
\end{algorithm}

\vspace{-0.15cm}
\subsection{Tri-prompt Dual-attention Fusion (TDF) Module}


In-context learning is designed to empower networks to adaptively optimize for various contexts by employing conditions or prompts that encapsulate external factors such as scene, task, or modality. Unlike the resource-intensive visual prompts or the rigid encodings of pretrained CLIP outputs or one-hot codes \cite{bahng2022exploring,liu2023clip,lin2023unifying}, which either burden computation and privacy or miss cross-client insights, we employ a triple-latent-prompt strategy. This approach integrates universal knowledge, data distribution, and annotation sparsity, effectively merging learnable parameters with predefined one-hot encodings for a more nuanced and efficient response. Concretely, all clients jointly maintain two learnable parameters: the universal knowledge prompt (UKP) $p_U\in \mathbb{R} ^{1\times H\times W}$ and the data distribution prompt (DDP) $p_D\in \mathbb{R} ^{N\times H\times W}$, with $N$ denoting the number of clients, and $H$ and $W$ respectively representing the height and width of the encoder's output feature $F$. The federation also provides $n$ one-hot annotation sparsity prompts (ASP) $p_S$, specifically set to 3 in this work, to discern between {sparse (e.g., point), medium (e.g., scribble), and dense (e.g., block) levels of sparsity, with box labels preprocessed into one of these three categories. ASP is designed to function as a switch to assist the model in perceiving the degree of annotation/supervision sparsity. It facilitates the coordination of training among multi-center models that are trained with supervision signals of varying sparsity levels.}

As illustrated in Fig. \ref{fig2}B, for the $i^{th}$ client, the local prompt $p_i$ is constructed by concatenating the shared UKP $p_U$, the locally selected DDP $p_{D,i}\in \mathbb{R} ^{1\times H\times W}$ from the channel dimension, and the dimensionally expanded ASP $p_{S,i}\in \mathbb{N}^{3\times H\times W}$, denoted as $p_i=\mathrm{cat}(p_U,p_{D,i},p_{S,i})$. 
{As suggested by previous works \cite{jia2022visual,wang2022personalizing,UniSeg2023}, directly adding or concatenating features with prompts may result in suboptimal performance. Rather, employing designs such as convolutional layers or some forms of attention mechanisms to effectively fuse and enhance both elements is more favorable.}
Given an encoded feature $F$, it is concatenated with $p_i$ and then passed through two convolutional blocks to obtain a preliminary fused feature $\hat{F}$. To further fuse and enhance features, we follow the typical dual-attention mechanism \cite{fu2019dual} in both spatial and channel dimensions. With $\otimes$ denoting the matrix multiplication operation, the spatial attention value $s_{ji}$ measuring the impact of the $i^{th}$ location on the $j^{th}$ location is calculated as 
\begin{equation}
    s_{ji}=\frac{\exp \left(Q^{\mathrm{T}}_{i} \otimes K_{j}\right)}{\sum_{i=1}^N \exp \left(Q^{\mathrm{T}}_{i} \otimes K_{j}\right)},
\end{equation}
and the channel attention value $c_{ji}$ measuring the impact of the $i^{th}$ channel on the $j^{th}$ channel is calculated as 
\begin{equation}
    c_{ji}=\frac{\exp \left(K'_{i} \otimes Q'^{\mathrm{T}}_{j}  \right)}{\sum_{i=1}^C \exp \left(K'_{i} \otimes Q'^{\mathrm{T}}_{j}\right)}.
\end{equation}
Then the fused feature $\tilde{F}$ is derived from summing the spatial enhanced feature $F_s$ and the channel enhanced feature $F_c$ which are respectively obtained by multiplying the matrix $V$ with the transpose of the spatial attention matrix $S$ and multiplying the transpose of the channel attention matrix $C$ with the matrix $V'$ ($V$ and $V'$ both are reshaped forms of $\hat{F}$, see Fig. \ref{fig2}B). 
The ultimate context-injected feature $\smash{\stackrel{*}{F}}$ is then formed by concatenating $F$ and $\tilde{F}$ using skip connection,
\begin{equation}
    {\tilde{F}=(\gamma_s \cdot F_s+\hat{F})+(\gamma_c \cdot F_c+\hat{F}),}
\end{equation}
\begin{equation}
    \stackrel{*}{F}=\mathrm{cat}(F,\tilde{F}),
\end{equation}
where $\gamma_s$ and $\gamma_c$ are two learnable parameters to balance the influence of the two types of enhanced features. Moreover, an MLP block featuring two linear layers is utilized to reduce the dimensionality of $\smash{\stackrel{*}{F}}$ and combine it with the features prior to the decoder's segmentation head via a Hadamard product, thereby enhancing its capacity for contextual adjustment.

\subsection{PDLA Mechanism and WSS Objective}
\label{subsec:PDLA}
In WSS, an effective strategy is to generate reliable pseudo-proposals for unlabeled areas. To reduce local computational overhead, we forgo the multi-level tree affinity approach \cite{lin2023unifying}.
{An alternative approach involves self-training using the model's own predictions. However, since weak labels typically lack direct supervision at boundary areas, models are prone to being constrained by their own predictions, especially in uncertain regions, which may lead to update stagnation or noise accumulation \cite{zhang2021adaptive,luo2022scribble}. To mitigate these issues, we adopt a dual-decoder strategy wherein an auxiliary branch incorporates beneficial information from other centers. This decoder provides predictions from a different or slightly different perspective. Predictions from the main decoder and the auxiliary decoder are then ensembled to yield pseudo labels of higher quality and more diversity.}
In the PDLA mechanism, parameter acquisition for both decoders involves two stages: aggregation or selection at the server side, and learnable aggregation at the local end.

As shown in Fig. \ref{fig2}A, at round $t$, the server collects the main decoder parameters $\{{\phi}_i^t\}^N_{i=1}$ and local prompts $\{p_i^t\}^N_{i=1}$ from all clients $\{c_i\}^N_{i=1}$, and extracts UKP $p_U^t$ and DDPs $\{p_{D,i}^t\}^N_{i=1}$. Then the inter-client similarity/affinity matrix $A$ is calculated via the cosine similarity
\begin{equation}
    \mathrm{sim}(\text{c}_i,\text{c}_j)=\frac{\mathrm{cat}(p_U^t,p_{D,i}^t)\cdot \mathrm{cat}(p_U^t,p_{D,j}^t)}{\|\mathrm{cat}(p_U^t,p_{D,i}^t)\|\cdot\|\mathrm{cat}(p_U^t,p_{D,j}^t)\|},
\end{equation}
\begin{equation}
    a_{ij}=\mathrm{ReLu}(\mathrm{sim}(\text{c}_i,\text{c}_j)),
\end{equation}
where $a_{ij}$ denotes similarity between the $i^{th}$ and $j^{th}$ clients and $\mathrm{ReLu}(x)=\max(0,x)$ is utilized to eliminate potential conflicts arising from dissimilar client models. 
In FedLPPA, four optional server-side auxiliary decoder parameter selection strategies are provided. {\textbf{Random} randomly allocates the collected $\{{\phi}_i^t\}^N_{i=1}$ to a different client as $\{\bar{\phi}_{G,i}^t\}^N_{i=1}$, whereas \textbf{Fixed Order} distributes them in a predefined cyclical sequence.
The \textbf{Highest Prompt Similarity} strategy selects ${\phi}_i^t$ corresponding to the highest $a_{ij}$ when $j\neq i$, while the \textbf{Prompt Similarity Aggregated} (default) strategy aggregates $\{{\phi}_i^t\}^N_{i=1}$ through the similarity matrix $A$,} 
\begin{equation}
    \bar{\phi}_{G,i}^{t}=\sum_{j=1}^N{\frac{a_{ij}}{\sum_{j=1}^N a_{ij}}}\phi _{i}^{t}.
\end{equation}

{To stabilize the training of the dual-decoder, to avoid unsuitable personalization due to coarse-grained aggregation, and to facilitate fine-grained and personalized knowledge transfer, we employ a Learnable Aggregation (LA) approach akin to residual learning. This approach involves learning an aggregation weight matrix for each decoder at the client end, which is then used to personalize the decoder parameters on an element/parameter-wise basis.} LA is formulated as
\begin{equation}
    \hat{\phi}_{i}^{t}:=\phi^{t-1}_i+\left(\phi_G^{t-1}-\phi^{t-1}_i\right) \odot W_{M,i},
\end{equation}
\begin{equation}
    \hat{\bar{\phi}}_{i}^{t}:=\bar{\phi}^{t-1}_i+\left(\bar{\phi}_{G,i}^{t-1}-\bar{\phi}^{t-1}_i\right) \odot W_{A,i},
\end{equation}
where $W_{M,i}$ and $W_{A,i}$ are learnable weight matrices, and $\sigma(w)=\max (0, \min (1, w)), \forall w \in \{W_{M,i},W_{A,i}\}$ is utilized to ensure the weight matrices' elements are within the range of $[0,1]$. $W_{M,i}$ and $W_{A,i}$ are initialized as {all-ones} matrices, and then iteratively get updated during training on local data 
\begin{equation}
    W_{M,i} \leftarrow W_{M,i}-\eta \nabla_{W_{M,i}} \mathcal{L_\text{wss}}\left(\hat{\phi}_{i}^{t}, D_i; \phi_G^{t-1}\right),
\end{equation}  
\begin{equation}
    W_{A,i} \leftarrow W_{A,i}-\eta \nabla_{W_{A,i}} \mathcal{L_\text{wss}}\left(\hat{\bar{\phi}}_{i}^{t}, D_i; \bar{\phi}_{G,i}^{t-1}\right),
\end{equation} 
where $\eta$ is the learning rate. During the update process of $W_{M,i}$ and $W_{A,i}$, Eqs. (10-11) and (12-13) are alternately executed for the updates of $(\hat{\phi}_{i}^{t},\hat{\bar{\phi}}_{i}^{t})$ and $(W_{M,i},W_{A,i})$. Upon convergence, the initial parameters $(\hat{\phi}_{i}^{t},\hat{\bar{\phi}}_{i}^{t})$ for the dual-decoder in each federation round are obtained, followed by the training of the local model using Eq. (1). Note that when $t>2$, only a few iterations are conducted to acquire $(W_{M,i},W_{A,i})$. 

For each client's WSS objective $\mathcal{L_\text{wss}}$, it is composed of direct supervision $\mathcal{L_\text{sup}}$ from sparse labels $y_k$ (employing the conventional partial cross-entropy \cite{tang2018normalized}) and $\mathcal{L_\text{pse}}$ supervised by pseudo-proposals $\tilde{y}_k$. And $\tilde{y}_k$ are constructed from the stochastic mixture of outputs from the dual decoders,
\begin{equation}
    \tilde{y}_k = \arg\max[\lambda_\text{m} \cdot p_{M,k} + (1-\lambda_\text{m}) \cdot p_{A,k}],
\end{equation}
where $p_{M,k}$ and $p_{A,k}$ respectively denote the $k^{th}$ sample's prediction probability from the main decoder and that from the auxiliary one, and $\lambda_\text{m} = \mathrm{Random}(0.7,1)$ is a hyperparameter used to randomly modulate the influence of the two decoders, while granting the main decoder a higher weight. With a trade-off parameter $\lambda$, $\mathcal{L_\text{wss}}$ is formulated as
\begin{equation}
\label{eq15}
    \mathcal{L_\text{wss}} = \mathcal{L_\text{sup}} + \lambda\mathcal{L_\text{pse}},
\end{equation}
\begin{equation}
    \mathcal{L_\text{sup}} = 0.5 \times \mathcal{L_\text{pCE}}(p_{M,k}, y_k) + 0.5 \times \mathcal{L_\text{pCE}}(p_{A,k}, y_k),
\end{equation}
\begin{equation}
    \mathcal{L_\text{pse}} = 0.5\times\mathcal{L_\text{Dice}}(p_{M,k}, \tilde{y}_k) +0.5\times\mathcal{L_\text{Dice}}(p_{A,k}, \tilde{y}_k).
\end{equation}

\subsection{Preprocessing Operations for Bounding Box Labels}

\begin{figure}[t]
    \centering
    \centerline{\includegraphics[width=0.97\columnwidth]{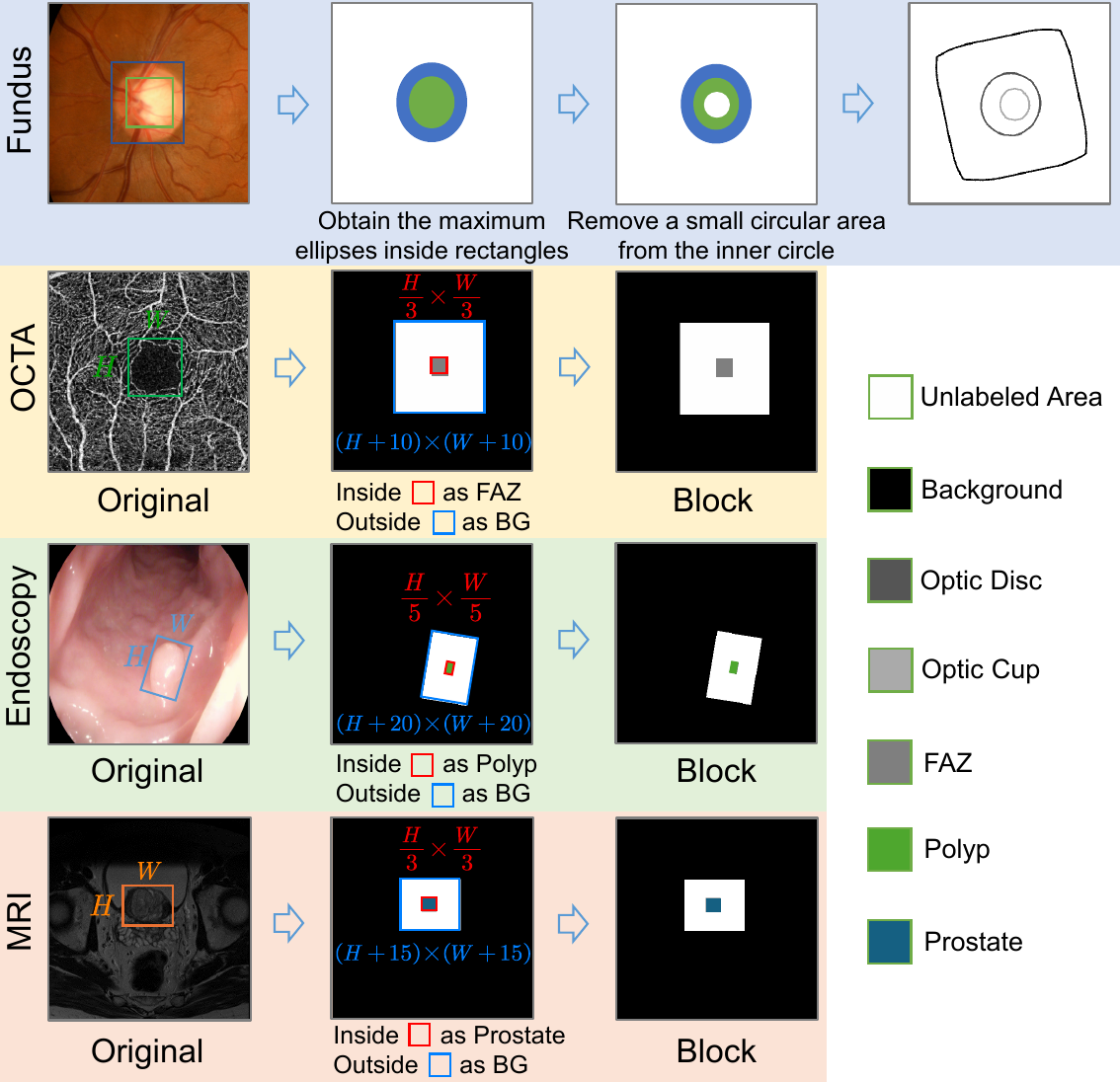}}
    \vspace{-0.1cm}
    \caption{Preprocessing operations for box annotations.}
    \label{fig3}
    \vspace{-0.4cm}
\end{figure}

As stated in Sec. \ref{subsec:wss}, box annotations do not provide direct pixel-level supervision but rather serve as indicators of position and range. Hence, we transform them into alternative types of sparse labels for paradigm alignment. In Fig. \ref{fig3}, we present the preprocessing steps for the box annotations of the {four} modalities investigated in this paper: {fundus image, OCTA image, MRI and endoscopy image (the first three employ conventional boxes, while the latter utilizes rotated ones). After conversion, their ASP $p_{S,i}$ is set to match: medium sparsity for fundus images (converted to scribbles) and dense sparsity for the other three (converted to blocks). These preprocessing steps are specific to our experiments. Other strategies, like using SAM-Med2D for segmentation (boxes as prompts) \cite{cheng2023sam} followed by moderate erosions to decrease label noise, are also viable.}

\vspace{-0.1cm}
\section{Experiments}
\subsection{Datasets and Preprocessing}
\label{dataset}
To verify the effectiveness of FedLPPA, we evaluate it on {four segmentation tasks involving distinct modalities of medical imaging: segmenting optic disc/cup (OD/OC) from fundus images, segmenting foveal avascular zone (FAZ) from OCT angiography (OCTA) images, segmenting polyps from endoscopy images, and segmenting prostate from MRI. As stated in Table \ref{table1}, all fundus, OCTA, endoscopy and MRI are acquired from various public datasets, possessing distinct data distributions even of the same modality. For fundus images, we follow Liu et al. \cite{liu2021feddg} to crop the images around the OD area and resize them to 384$\times$384. For OCTA and endoscopy images, we directly resize them to 256$\times$256 and 384$\times$384. For MRI, we utilize the multi-center dataset constructed by Liu et al. \cite{liu2020ms} for prostate MRI segmentation, dividing the cases into training and testing sets at a predetermined ratio and slicing them (slices not containing the prostate are excluded). The slices are then resized to 384$\times$384.} Data preprocessing includes normalizing all image intensities to [0,1], while data augmentation comprises random horizontal and vertical flips as well as random rotations (spanning from -45° to 45°).

Regarding the various sparse annotations employed, they are synthesized using automated algorithms based on the original full masks provided by each dataset, primarily implemented through OpenCV2, SimpleITK, and Scikit-image. For \textbf{points}, we obtain the maximal inscribed rectangle from the mask and reduce it by a certain proportion. We then take the midpoints of the four sides of the shrunken rectangle and multiply them by a Gaussian kernel to simulate human brushstrokes. The acquisition of \textbf{scribbles} primarily relies on morphological erosion and skeletonization processes, while \textbf{another style of scribble} undergoes additional elastic transformations and random erasures. \textbf{Block annotations} are derived via morphological erosion. \textbf{Standard bounding boxes} are derived from the masks' encompassing rectangles, while \textbf{rotated ones} come from their minimal enclosing rectangles.

\vspace{-0.1cm}

\subsection{Implementation Details}

All compared FL methods and the proposed FedLPPA are implemented with PyTorch using NVIDIA A100 GPUs. We employ the vanilla UNet as the model architecture, with the number of channels progressively increasing from 16 to 256 from top to bottom. In FedLPPA, the auxiliary decoder shares the same architecture as the vanilla UNet decoder, and an MLP block is added to the main decoder. We optimize parameters using the AdamW optimizer with an initial learning rate of 1 $\times$ 10$^{-2}$, and dynamically adjust the learning rate using a polynomial strategy with a power of 0.9. 
The trade-off parameter $\lambda$ utilized in $\mathcal{L_\text{wss}}$ (Eq. \ref{eq15}) and the batch size are respectively set to 0.5 and 12. We train all FL methods for 500 federation rounds to ensure a fair performance comparison. In each round of local updates, all clients undergo training for ten iterations, and in FedLPPA, two additional iterations are utilized to update $W_{M,i}$ and $W_{A,i}$ when $t>2$.


\begin{table}[t]
    \caption{
        {Details of the datasets utilized in our experiments. Scribble$^2$ refers to scribbles of a different style generated using another algorithm.}}
    \label{table1}
    \setlength{\tabcolsep}{0.7mm}
    \vspace{-0.1cm}
    \centering
    \resizebox{\columnwidth}{!}{
    \begin{tabular}{cccccc}
        \specialrule{0.12em}{0pt}{0.5pt}
        Modality                & Site & Original Dataset & \begin{tabular}[c]{@{}c@{}}Train-Test \\ Sample Size\end{tabular} & Annotation Type & \begin{tabular}[c]{@{}c@{}}Processed \\ Resolution {[}px{]}\end{tabular} \\
        \specialrule{0.05em}{0.5pt}{0.5pt}
        \multirow{5}{*}{Fundus} & A    & Drishti-GS1\cite{sivaswamy2015comprehensive}        & 50:51                                                             & Scribble         & \multirow{5}{*}{384$\times$384}                                            \\
                                & B    & RIM-ONE-r3\cite{fumero2011rim}       & 99:60                                                             & Scribble$^2$        &                                                                        \\
                                & C    & REFUGE-train\cite{orlando2020refuge}     & 320:80                                                            & Bounding box     &                                                                          \\
                                & D    & REFUGE-val\cite{orlando2020refuge}       & 320:80                                                            & Point            &                                                                          \\
                                & E    & GAMMA\cite{wu2023gamma}           & 100:100                                                           & Block            &                                                                          \\
        \specialrule{0.05em}{0.5pt}{0.5pt}
        \multirow{5}{*}{OCTA}   & A    & FAZID\cite{agarwal2020foveal}          & 244:60                                                            & Scribble         & \multirow{5}{*}{256$\times$256}                                                 \\
                                & B    & OCTA500-3M\cite{li2020ipn}       & 150:50                                                            & Point            &                                                                         \\
                                & C    & OCTA500-6M\cite{li2020ipn}       & 200:100                                                           & Block            &                                                                        \\
                                & D    & OCTA-25K (3x3)\cite{wang2021deep} & 708:304                                                           & Bounding box     &                                                                        \\
                                & E    & ROSE\cite{ma2020rose}             & 30:9                                                              & Scribble$^2$        &                                                                        \\
                                \specialrule{0.05em}{0.5pt}{0.5pt}
                                \multirow{4}{*}{Endoscopy}   & A    & CVC-ClinicDB\cite{bernal2015wm}
                                & 550:62                                                            & Point         & \multirow{4}{*}{384$\times$384}                                                 \\
                                & B    & CVC-ColonDB\cite{vazquez2017benchmark}
                                & 327:63                                                            & Scribble            &                                                                         \\
                                & C    & ETIS-Larib\cite{silva2014toward}                                & 170:26                                                           & Rotated box            &                                                                        \\
                                & D    & Kvasir-SEG\cite{jha2020kvasir}
                                & 900:100                                                           & Block     &                                                                        \\
                                \specialrule{0.05em}{0.5pt}{0.5pt}

                                \multirow{6}{*}{MRI}   & A    &  NCI-ISBI 2013 (Siemens)\cite{bl2013nciisbi}
                                & 340:81                                                            & Block         & \multirow{6}{*}{384$\times$384}                                                 \\
                                & B    & NCI-ISBI 2013 (Philips)\cite{bl2013nciisbi}
                                & 297:87                                                            & Point            &                                                                         \\
                                & C    & I2CVB\cite{lemaitre2015computer}                                & 372:96                                                           & Scribble            &                                                                        \\
                                & D    & PROMISE12 (Siemens-UCL)\cite{litjens2014evaluation}
                                & 134:41                                                           & Point     &   \\  
                                & E    & PROMISE12 (GE-BIDMC)\cite{litjens2014evaluation}
                                & 193:68                                                           & Scribble$^2$     &  \\ 
                                & F    & PROMISE12 (Siemens-HK)\cite{litjens2014evaluation}
                                & 122:36                                                           & Bounding box     &                                                                       \\
                                \specialrule{0.12em}{0pt}{0pt}
                                
        \end{tabular}
        }
\vspace{-0.4cm}  
\end{table}

\subsection{Comparisons with State-of-the-art}
\label{sota exp}

\begin{table*}[!h]
    \vspace{-0.45cm}
    \caption{
        Performance comparisons of different FL methods as well as different local and centralized training settings on ODOC segmentation. The best results are highlighted in bold and the second-best ones are underlined.}
    \label{table2}
    \setlength{\tabcolsep}{0.35mm}
    \centering
    \resizebox{\textwidth}{!}{
        \begin{tabular}{l|cccccc|cccccc|c|cccccc|cccccc|c}
            \specialrule{0.12em}{0pt}{0pt}
            & \multicolumn{6}{c|}{DSC (OD) $\uparrow$}                                                    & \multicolumn{6}{c|}{DSC (OC) $\uparrow$}    & \multirow{2}{*}{Overall}                                                 &  \multicolumn{6}{c|}{HD95 (OD) $\downarrow$}                                                    & \multicolumn{6}{c|}{HD95 (OC)$\downarrow$}   & \multirow{2}{*}{Overall}     \\
            \cline{1-13} \cline{15-26}
            Methods & Site A          & Site B           & Site C          & Site D           & Site E           & \textbf{Avg.}  & Site A           & Site B           & Site C          & Site D           & Site E           & \textbf{Avg.}  & & Site A          & Site B           & Site C          & Site D           & Site E           & \textbf{Avg.}  & Site A           & Site B           & Site C          & Site D           & Site E           & \textbf{Avg.}  \\
\hline
FedAvg\cite{mcmahan2017communication}         & 92.58 & 83.90 & 94.67 & 93.36  & 88.92  & 90.68 & 83.16 & 69.91 & {85.35} & 82.80 & 86.76 & 81.60 & 86.14 & 13.17  & 24.10 & 9.28    & 20.80  & 24.59 & 18.39 & 18.56 & 23.98 & {11.19}  & 22.42 & 16.80 & 18.59 &  18.49 \\
FedProx\cite{li2020federated}        & 95.19 & 82.14 & \underline{95.03} & 88.28 & {91.06}  & 90.34 & 82.18 & 70.14 & 84.64 & 81.33 & {87.19}  & 81.10  & 85.72   & \underline{8.38}  & {15.72} & 10.64  & 16.02  & 20.36 & 14.22 & {14.17}  & 12.82  & 13.36 & 13.91 & 16.10 & 14.07 & 14.15  \\
FT\cite{wang2019federated}              & {95.97} & 91.09  & 94.83 & 93.34  & 90.10  & 93.06 & {85.02} & 80.76 & 82.51 & 82.95  & 86.88 & {83.62}  &   88.34  &  10.07  & 22.26 & \underline{8.51}    & 40.12  & 20.36 & 20.26 & 19.96 & 22.24 & 11.68  & 31.60 & 14.59 & 20.02 & 20.14   \\
FedBN\cite{li2020fedbn}           & 95.88 & \textbf{92.66}  & 94.80 & {95.06}  & 90.72 & 93.82 & 84.54 & \underline{82.09} & 83.79 & 86.48  & 79.38 & 83.25  & 88.54    &  8.78   & 18.06 & 19.54  & {10.72}  & 17.68 & 14.96 & 15.70  & \underline{12.08}  & 17.24 & 9.96  & {11.96} & 13.39 &  14.17 \\
FedAP\cite{lu2022personalized}           & 95.81 & \underline{92.24}  & \textbf{95.45} & \underline{95.17}  & 90.73 & {93.88} & 83.97 & 79.89 & 83.55 & {86.78}  & 83.47 & 83.53 &  88.71  & 9.19   & 21.59 & {8.96}   & \textbf{5.77}    & {16.80} & \underline{12.46} & 15.69  & 14.68 & 12.66 & \underline{7.05}   & 14.49 & {12.91} &  12.69 \\
FedRep\cite{collins2021exploiting}          & 95.28 & 88.10  & 92.68 & 92.34  & 88.75 & 91.43 & 82.47 & 76.22 & 82.14 & 83.04  & 81.29 & 81.03 & 86.23 & 17.39 & 21.59 & 23.56  & 54.10  & 31.00 & 29.53 & 27.61 & 18.90 & 17.89 & 50.60 & 17.71 & 26.54  & 28.03  \\
MetaFed\cite{chen2023metafed}       & 95.88 & 91.97  & 89.78 & 93.51  & 86.44 & 91.51 & 83.69 & 81.86 & 84.72 & 85.04  & 82.24 & 83.51  & 87.51  & 18.41 & 27.92 & 114.22 & 31.53  & 19.21 & 42.26 & 18.93 & 17.45  & 21.63 & 8.46  & 15.22 & 16.34 &  29.30 \\
FedLC\cite{wang2022personalizing}  & 95.77 & 88.86  & 93.04 & 94.23  & 94.03  & 93.19 & 83.69 & 74.90 & 86.63 & 86.17  & \underline{88.96} & 84.07  &  88.63   & 11.56   & 18.36 & 10.42    & 9.16    & 14.82  & 12.86  & 21.71 & 18.57 & 10.09   & 10.28   & 14.38  & 15.01  & 13.94  \\
FedALA\cite{zhang2023fedala} & 95.02 & 87.86  & 94.42 & 92.04  & 94.15  & 92.70 & 83.85 & 76.22 & \textbf{88.00} & 86.02  & \textbf{89.35} & 84.69  &  88.69  & 17.51   & 16.76 & 9.57    & 46.05    & \underline{11.02}  & 20.18  & 19.76 & 14.10 & \underline{8.89}   & 16.29   & \textbf{7.24}  & 13.26 &   16.72  \\
FedICRA\cite{lin2023unifying}  & \underline{96.18} & 91.41  & \underline{95.03} & \textbf{95.34}  & \underline{95.18}  & \underline{94.63} & \underline{85.17} & 79.26 & \underline{87.43} & \textbf{88.38}  & 88.87 & \underline{85.82}   &  \underline{90.23}   & 8.66   & 35.53 & \textbf{7.95}    & \underline{7.46}    & \textbf{10.24}  & 13.97  & \underline{14.61} & 13.82 & 8.97   & 7.50   & 9.46  & \underline{10.87}  &   \underline{12.42}  \\
FedLPPA (Ours) & \textbf{96.67} & \underline{92.24}  & 94.72 & 94.54  & \textbf{95.39}  & \textbf{94.71} & \textbf{86.89} & \textbf{83.35} & 87.22 & \underline{87.88}  & 88.71 & \textbf{86.81}  &   \textbf{90.76}  & \textbf{7.46}   & \textbf{9.72} & 8.66    & 8.84    & 13.03  & \textbf{9.54}  & \textbf{12.62} & \textbf{10.53} & \textbf{8.76}   & \textbf{6.38}   & \underline{8.19}  & \textbf{9.30}  &  \textbf{9.42} \\
\hline

LT (Weak)\cellcolor{gray!20}      & 94.83\cellcolor{gray!20}& 90.65\cellcolor{gray!20} & 89.31\cellcolor{gray!20} & 76.38\cellcolor{gray!20} & 86.11\cellcolor{gray!20} & 87.45\cellcolor{gray!20} & 84.50\cellcolor{gray!20} & 79.31\cellcolor{gray!20} & 84.11\cellcolor{gray!20} & 77.02\cellcolor{gray!20}  & 81.38\cellcolor{gray!20} & 81.26\cellcolor{gray!20} &  84.36\cellcolor{gray!20} & 26.91\cellcolor{gray!20} & 44.05\cellcolor{gray!20} & 74.35\cellcolor{gray!20}  & 151.71\cellcolor{gray!20} & 19.53\cellcolor{gray!20} & 63.31\cellcolor{gray!20} & 16.51\cellcolor{gray!20}  & 35.98\cellcolor{gray!20} & 24.93\cellcolor{gray!20} & 19.27\cellcolor{gray!20} & 16.45\cellcolor{gray!20} & 22.62\cellcolor{gray!20} &  42.97\cellcolor{gray!20} \\

CT (Weak)\cellcolor{gray!20}       & 95.45\cellcolor{gray!20}  & 91.41\cellcolor{gray!20}   & 95.45\cellcolor{gray!20}  & 91.92\cellcolor{gray!20}   & 91.21\cellcolor{gray!20}   & 93.09\cellcolor{gray!20}  & 84.25\cellcolor{gray!20}  & 78.94\cellcolor{gray!20}  & 84.63\cellcolor{gray!20}  & 85.68\cellcolor{gray!20}   & 87.08\cellcolor{gray!20}  & 84.12\cellcolor{gray!20}  &  88.60\cellcolor{gray!20}  & 9.32\cellcolor{gray!20}   & 12.25\cellcolor{gray!20}  & 7.29\cellcolor{gray!20}    & 11.97\cellcolor{gray!20}  & 14.52\cellcolor{gray!20} & 11.07\cellcolor{gray!20} & 15.55\cellcolor{gray!20}  & 13.79\cellcolor{gray!20}  & 10.73\cellcolor{gray!20}  & 9.83\cellcolor{gray!20}  & 12.15\cellcolor{gray!20} & 12.41\cellcolor{gray!20}  &  11.74\cellcolor{gray!20} \\
LT (Full)\cellcolor{gray!20}      & 96.28\cellcolor{gray!20}  & 96.21\cellcolor{gray!20}   & 95.21\cellcolor{gray!20}  & 95.56\cellcolor{gray!20}   & 94.33\cellcolor{gray!20}  & 95.52\cellcolor{gray!20}  & 87.94\cellcolor{gray!20}   & 81.72\cellcolor{gray!20}  & 87.22\cellcolor{gray!20}  & 88.60\cellcolor{gray!20}   & 86.53\cellcolor{gray!20}  & 86.40\cellcolor{gray!20}  &  90.96\cellcolor{gray!20} & 9.27\cellcolor{gray!20}  & 6.26\cellcolor{gray!20}   & 7.55\cellcolor{gray!20}    & 5.38\cellcolor{gray!20}    & 11.07\cellcolor{gray!20} & 7.91\cellcolor{gray!20}  & 13.34\cellcolor{gray!20}  & 12.62\cellcolor{gray!20}  & 9.18\cellcolor{gray!20}   & 6.30\cellcolor{gray!20}   & 11.51\cellcolor{gray!20} & 10.59\cellcolor{gray!20}  &  9.25\cellcolor{gray!20}  \\
CT (Full)\cellcolor{gray!20}      & 96.97\cellcolor{gray!20}  & 94.28\cellcolor{gray!20}   & 96.06\cellcolor{gray!20}  & 96.20\cellcolor{gray!20}  & 95.52\cellcolor{gray!20}   & 95.81\cellcolor{gray!20}  & 87.40\cellcolor{gray!20}  & 82.90\cellcolor{gray!20}   & 88.30\cellcolor{gray!20} & 89.17\cellcolor{gray!20}   & 89.25\cellcolor{gray!20}  & 87.40\cellcolor{gray!20}  &  91.61\cellcolor{gray!20} & 6.40\cellcolor{gray!20}   & 8.95\cellcolor{gray!20}   & 6.13\cellcolor{gray!20}    & 4.62\cellcolor{gray!20}    & 8.92\cellcolor{gray!20}  & 7.00\cellcolor{gray!20}     & 12.31\cellcolor{gray!20}  & 11.28\cellcolor{gray!20}  & 8.47\cellcolor{gray!20}   & 5.74\cellcolor{gray!20}   & 8.48\cellcolor{gray!20}  & 9.26\cellcolor{gray!20}  &  8.13\cellcolor{gray!20} \\
\specialrule{0.12em}{0pt}{0pt}

\end{tabular}
}
\vspace{-0.1cm}
\end{table*}

\begin{figure*}[t]
    \centering
    \includegraphics[width=\textwidth]{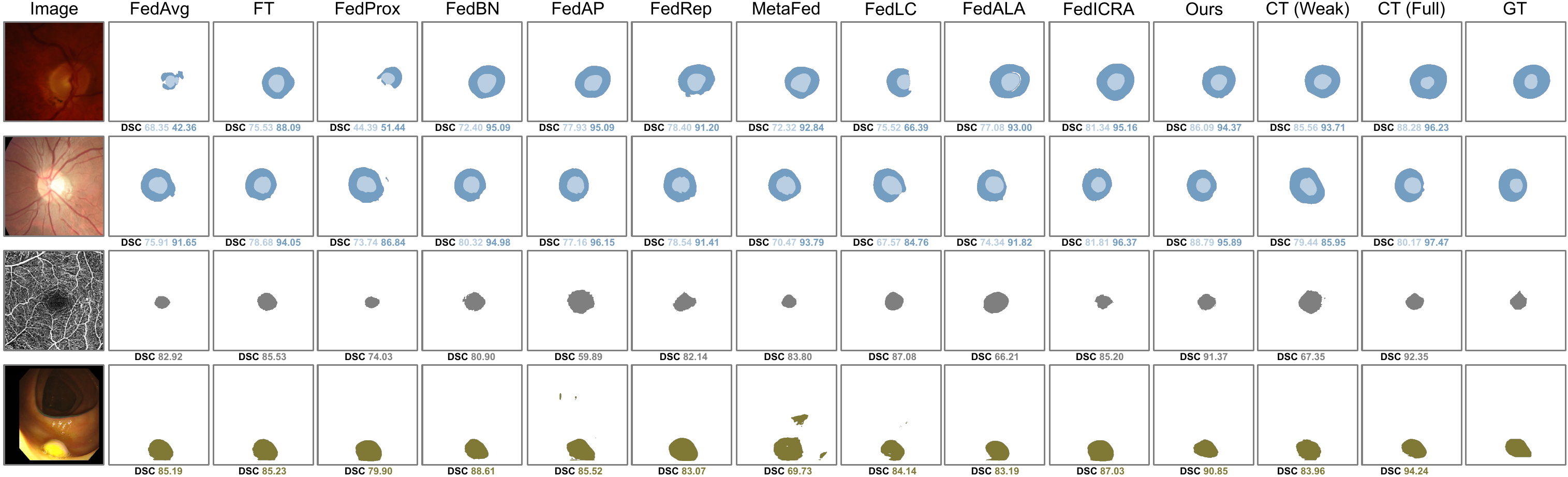}
    \vspace{-0.6cm}
    \caption{Visualization results from FedLPPA and other SOTA methods (\textbf{CT} denotes centralized training, with \textbf{Weak} and \textbf{Full} respectively indicating the use of sparse annotations and full masks for model training).} 
    \label{fig4}
    \vspace{-0.4cm}
\end{figure*}

We benchmark FedLPPA against several representative FL frameworks, encompassing both commonly employed gFL methods such as FedAvg \cite{mcmahan2017communication} and FedProx \cite{li2020federated}, and state-of-the-art (SOTA) pFL methods including FT \cite{wang2019federated}, FedBN \cite{li2020fedbn}, FedAP \cite{lu2022personalized}, FedRep \cite{collins2021exploiting}, MetaFed \cite{chen2023metafed}, FedLC \cite{wang2022personalizing}, FedALA \cite{zhang2023fedala} and FedICRA \cite{lin2023unifying}, etc. 
FedAvg computes a weighted average of the parameters from each local model, based on the sample size, to obtain a singular global model. FedProx is an extension of FedAvg, incorporating a proximal term into optimization to address system heterogeneity, thereby enhancing stability and performance in heterogeneous environments. In pFL methods, FT represents FedAvg with site fine-tuning. FedBN and FedRep respectively treat all batch normalization (BN) layers and the task head (e.g., the final convolutional layer) as the personalized components of the model. FedAP aggregates through BN layer statistics to assess similarities across sites, while preserving individualized parameters of each site's BN layers. MetaFed achieves personalization for each site through static (fixed order) cyclic knowledge distillation. FedALA employs an adaptive local aggregation module to facilitate knowledge transfer between the global model and local ones, thereby realizing personalization. FedLC achieves personalization by employing local calibrations that leverage inconsistencies at both the feature- and prediction levels. FedICRA accommodates the same setting as this paper by utilizing one-hot prompts and novel objective functions for WSS.
Considering that some of the above methods are initially designed for classification, we endeavor to preserve their original design tenets while adapting them for segmentation tasks. All comparison methods, being consistent with FedLPPA, are trained under a WSS setting on the same datasets described in Sec. \ref{dataset} . Additionally, comparisons are made with some baselines and ideal setups, including local training (LT) and centralized training (CT) with weak and full labels. All methods are evaluated using two metrics: the Dice Similarity Coefficient (DSC) and the 95\% Hausdorff Distance (HD95 [px]).

Table \ref{table2} presents the quantitative results for the ODOC segmentation task. The \textbf{Overall} columns demonstrate that compared to LT, all FL methods enhance the overall segmentation performance across all sites. Sites C and E achieve significant performance improvements after participating in FL, as their diverse local data distributions make it challenging for them to train sufficiently robust models using only local data. Site D also experiences a notable performance increase after joining FL, due to its use of point annotations, where the supervisory signals are insufficient for training an adequate model. Due to the high data heterogeneity among sites in the fundus scenario, as depicted in Fig. \ref{fig:tsne&cor}A, pFL methods, which train personalized models for different sites, typically achieve higher performance than gFL methods. FedAP and FedICRA achieve notable performance, while FedLPPA further enhances it. All clients benefit within our framework, showing significant superiority over CL (weak) and closely approaching CL (full) in the overall performance.

\begin{table}[tbp]
    \caption{
        Performance comparisons of different FL methods as well as different local and centralized training settings on FAZ segmentation.}
    \label{table3}
    \setlength{\tabcolsep}{0.4mm}
    \vspace{-0.15cm}
    \centering
    \resizebox{\columnwidth}{!}{
        \begin{tabular}{l|cccccc|cccccc}
            \specialrule{0.12em}{0pt}{0pt}
            & \multicolumn{6}{c|}{DSC (FAZ) $\uparrow$}                                                    &  \multicolumn{6}{c}{HD95 (FAZ) $\downarrow$}                                                    \\
\hline
            Methods & Site A          & Site B           & Site C          & Site D           & Site E           & \textbf{Avg.}  & Site A           & Site B           & Site C          & Site D           & Site E           & \textbf{Avg.}  \\
\hline
FedAvg\cite{mcmahan2017communication}   & 76.56 & 90.65 & 77.64 & 88.85 & 86.55  & 84.05 & \underline{5.85}   & 8.28   & 22.73 & 12.45 & 21.59 & 14.18        \\
FedProx\cite{li2020federated}  & 79.93 & 92.11 & 83.26 & 91.68 & 90.71  & 87.54 & 10.88   & 6.79   & 9.13 & 7.71 & 7.19 & 8.34        \\
FT\cite{wang2019federated}   & {83.18}  & {91.96} & 78.58 & 88.75 & {88.10}  & {86.11} & 10.11 & 8.86  & 20.84 & 11.53 & 20.17 & 14.30             \\
FedBN\cite{li2020fedbn}    & 70.48 & 93.69 & 81.66 & 90.36 & 88.23 & 84.89 & 21.42 & 7.65  & 10.17 & 8.33 & 8.76  & 11.27    \\
FedAP\cite{lu2022personalized}   & 62.23 & 87.82 & 72.67 & 90.36 & 66.00  & 75.82 & 18.50 & {7.71}   & {14.51} & 8.30   & {18.60}  & {13.52}          \\
FedRep\cite{collins2021exploiting}   & 78.55 & 91.69 & {79.29} & {91.04} & 86.83 & 85.48 & 16.07 & 9.14  & 15.24 & {7.64}   & 19.92 & 13.60         \\
MetaFed\cite{chen2023metafed}   & 70.99 & 86.22 & 74.15 & 88.77 & 74.62 & 78.95 & 9.18  & 10.37  & 24.61 & 13.91 & 39.61 & 19.54     \\
FedLC\cite{wang2022personalizing}  &81.57&92.65&83.71&92.70&90.76&88.28&10.44&6.77&8.40&6.87&7.74&8.04 \\
FedALA\cite{zhang2023fedala}  &77.57&90.44&82.34&91.97&90.48&86.56&9.47&7.00&8.92&7.35&6.80&7.90  \\
FedICRA\cite{lin2023unifying}  &\underline{85.95}&\underline{96.94}&\underline{89.44}&\underline{93.68}&\underline{94.79}&\underline{92.16}&{7.47}&\underline{4.61}&\underline{8.05}&\underline{6.09}&\textbf{3.79}&\underline{6.00} \\
FedLPPA (Ours)&\textbf{89.40}&\textbf{97.36}&\textbf{91.00}&\textbf{94.24}&\textbf{95.56}&\textbf{93.51}&\textbf{5.35}&\textbf{4.20}&\textbf{7.22}&\textbf{5.45}&\underline{4.28}&\textbf{5.30} \\
\hline

LT (Weak)\cellcolor{gray!20}  & 73.74\cellcolor{gray!20} & 91.90\cellcolor{gray!20} & 79.03\cellcolor{gray!20} & 85.78\cellcolor{gray!20} & 79.89\cellcolor{gray!20} & 82.07\cellcolor{gray!20} & 30.94\cellcolor{gray!20} & 42.49\cellcolor{gray!20} & 20.81\cellcolor{gray!20} & 10.87\cellcolor{gray!20}  & 72.08\cellcolor{gray!20} & 35.44\cellcolor{gray!20}       \\

CT (Weak)\cellcolor{gray!20}   & 74.93\cellcolor{gray!20} & 89.24\cellcolor{gray!20} & 76.64\cellcolor{gray!20} & 89.25\cellcolor{gray!20} & 86.50\cellcolor{gray!20}  & 83.31\cellcolor{gray!20} & 8.38\cellcolor{gray!20}   & 7.54\cellcolor{gray!20}   & 10.46\cellcolor{gray!20} & 8.87\cellcolor{gray!20}   & 8.28\cellcolor{gray!20}   & 8.7\cellcolor{gray!20}         \\
LT (Full)\cellcolor{gray!20}  & 90.88\cellcolor{gray!20}  & 97.75\cellcolor{gray!20} & 89.22\cellcolor{gray!20} & 95.14\cellcolor{gray!20} & 95.11\cellcolor{gray!20}  & 93.62\cellcolor{gray!20} & 5.31\cellcolor{gray!20}   & 3.36\cellcolor{gray!20}   & 7.84\cellcolor{gray!20}  & 4.80\cellcolor{gray!20}   & 4.54\cellcolor{gray!20}   & 5.17\cellcolor{gray!20}        \\
CT (Full)\cellcolor{gray!20}     & 90.93\cellcolor{gray!20} & 97.23\cellcolor{gray!20} & 91.49\cellcolor{gray!20}  & 94.88\cellcolor{gray!20} & 95.38\cellcolor{gray!20}  & 93.98\cellcolor{gray!20} & 4.57\cellcolor{gray!20}   & 3.66\cellcolor{gray!20}   & 6.86\cellcolor{gray!20}  & 4.82\cellcolor{gray!20}   & 2.99\cellcolor{gray!20}   & 4.58\cellcolor{gray!20}     \\
\specialrule{0.12em}{0pt}{0pt}

\end{tabular}
}
\vspace{-0.5cm}
\end{table}

Tables \ref{table3} and \ref{table4} respectively tabulate the quantitative results on the FAZ and Polyp segmentation tasks. As illustrated in Figs. \ref{fig:tsne&cor}B and \ref{fig:tsne&cor}C, unlike the fundus scenario, the distribution variations in OCTA and endoscopy images across different clients are relatively minor. Some pFL methods (like FedBN, FedAP, FedRep, and MetaFed), due to inappropriate or coarse-grained personalization, even underperform typical gFL methods like FedAvg and FedProx. Benefiting from the proposed TDF module which enables in-context learning for adaptive adjustments to different scenarios, and the PDLA mechanism, FedLPPA can make fine-grained personalized adjustments. Further, FedLPPA enables each client model to perceive different scenarios in an in-context learning manner and make finer-grained personalized adjustments. As a result, it achieves superior performance in various data heterogeneity tasks. Notably, in the FAZ segmentation task, FedLPPA significantly outperforms other FL methods and comes very close to the performance under the CT (full) setting. It is also worth mentioning that in the Polyp segmentation scenario, FedLPPA not only greatly surpasses its competitors but also exceeds the CT (full) setting by 4.6\% in DSC, fully demonstrating the potential of WSS combined with FL. {Table \ref{table4-new} tabulates the quantitative results for the prostate segmentation task. It is evident that FedLPPA continues to achieve the best overall performance, significantly surpassing CT (weak) and exhibiting the smallest disparity with CT (full), compared to other representative gFL and pFL methods. Representative visualization results are provided in Fig. \ref{fig4} and Fig. \ref{fig4-new}.}

\begin{table}[t]
    \vspace{-0.45cm}
    \caption{
        Performance comparisons of different FL methods as well as different local and centralized training settings on polyp segmentation.}
    \label{table4}
    \setlength{\tabcolsep}{0.4mm}
    \vspace{-0.15cm}
    \centering
    \resizebox{\columnwidth}{!}{
        \begin{tabular}{l|ccccc|ccccc}
            \specialrule{0.12em}{0pt}{0pt}
            & \multicolumn{5}{c|}{DSC (Polyp) $\uparrow$}                                                    &  \multicolumn{5}{c}{HD95 (Polyp) $\downarrow$}                                                    \\
\hline
            Methods & Site A          & Site B           & Site C          & Site D                      & \textbf{Avg.}  & Site A           & Site B           & Site C          & Site D                     & \textbf{Avg.}  \\
\hline
FedAvg\cite{mcmahan2017communication}   &67.05&67.32&36.07&74.62&61.27&81.22&50.27&103.67&67.43&75.65        \\
FedProx\cite{li2020federated}  &64.36&75.28&39.47&74.09&63.30&70.87&33.68&93.61&65.65&65.95          \\
FT\cite{wang2019federated}   &69.84&67.62&36.64&74.44&62.14&92.13&51.68&92.53&63.30&74.91             \\
FedBN\cite{li2020fedbn}   &42.60&53.16&41.40&75.47&53.16&185.47&91.22&87.05&57.52&105.31         \\
FedAP\cite{lu2022personalized}   &47.65&58.22&26.26&72.94&51.27&179.96&98.69&90.24&59.31&107.05          \\
FedRep\cite{collins2021exploiting}  &63.96&74.96&5.28&76.22&55.10&130.52&24.46&121.19&62.16&84.58          \\
MetaFed\cite{chen2023metafed} &57.05&56.23&16.94&74.14&51.09&135.32&90.71&127.69&63.48&104.30       \\
FedLC\cite{wang2022personalizing} &76.22&\underline{77.03}&63.08&\underline{82.28}&\underline{74.65}&71.93&35.19&85.21&55.78&62.03  \\
FedALA\cite{zhang2023fedala} &74.38&76.21&53.12&79.96&70.92&78.82&40.12&100.21&70.79&72.48  \\
FedICRA\cite{lin2023unifying}  &\underline{77.46}&76.17&\underline{63.57}&80.83&74.51&\underline{47.03}&\underline{30.20}&\underline{67.24}&\underline{48.76}&\underline{48.31}\\
FedLPPA (Ours) &\textbf{84.38}&\textbf{87.65}&\textbf{67.38}&\textbf{84.56}&\textbf{80.99}&\textbf{29.30}&\textbf{14.59}&\textbf{62.98}&\textbf{43.51}&\textbf{37.60} \\
\hline

LT (Weak)\cellcolor{gray!20}     &46.65\cellcolor{gray!20}&44.09\cellcolor{gray!20}&19.74\cellcolor{gray!20}&77.75\cellcolor{gray!20}&47.06\cellcolor{gray!20}&183.56\cellcolor{gray!20}&152.43\cellcolor{gray!20}&94.28\cellcolor{gray!20}&54.44\cellcolor{gray!20}&121.18\cellcolor{gray!20}   \\

CT (Weak)\cellcolor{gray!20}    &57.12\cellcolor{gray!20}&60.77\cellcolor{gray!20}&29.93\cellcolor{gray!20}&76.85\cellcolor{gray!20}&56.17\cellcolor{gray!20}&78.65\cellcolor{gray!20}&63.18\cellcolor{gray!20}&82.18\cellcolor{gray!20}&52.26\cellcolor{gray!20}&69.07\cellcolor{gray!20}     \\
LT (Full)\cellcolor{gray!20}     &87.83\cellcolor{gray!20}&74.08\cellcolor{gray!20}&49.83\cellcolor{gray!20}&86.58\cellcolor{gray!20}&74.58\cellcolor{gray!20}&25.34\cellcolor{gray!20}&43.11\cellcolor{gray!20}&85.12\cellcolor{gray!20}&50.33\cellcolor{gray!20}&50.97\cellcolor{gray!20}     \\
CT (Full)\cellcolor{gray!20}     &82.72\cellcolor{gray!20}&79.79\cellcolor{gray!20}&59.01\cellcolor{gray!20}&83.78\cellcolor{gray!20}&76.32\cellcolor{gray!20}&28.30\cellcolor{gray!20}&22.00\cellcolor{gray!20}&65.96\cellcolor{gray!20}&50.59\cellcolor{gray!20}&41.72\cellcolor{gray!20}    \\
\specialrule{0.12em}{0pt}{0pt}

\end{tabular}
}
\vspace{-0.1cm}
\end{table}

\begin{table}[t]
    \vspace{-0.1cm}
    \caption{
        {Performance comparisons of different FL methods as well as different local and centralized training settings on prostate segmentation. Note: The word "Site" preceding the site letters is omitted to conserve table space.}}
    \label{table4-new}
    \setlength{\tabcolsep}{0.5mm}
    \vspace{-0.15cm}
    \centering
    \resizebox{\columnwidth}{!}{
        \begin{tabular}{l|ccccccc|ccccccc}
            \specialrule{0.12em}{0pt}{0pt}
            & \multicolumn{7}{c|}{DSC (Prostate) $\uparrow$}                                                    &  \multicolumn{7}{c}{HD95 (Prostate) $\downarrow$}                                                    \\
\hline
            Methods & A          & B           & C          & D      & E   & F              & \textbf{Avg.}  & A           & B           & C          & D    & E   & F                 & \textbf{Avg.}  \\
\hline
FedAvg\cite{mcmahan2017communication}   &83.83&81.93&78.16&80.79&66.47&82.12&78.88&10.33&28.49&31.29&\underline{10.77}&36.23&17.01&22.35        \\
FedProx\cite{li2020federated}  &83.03&82.76&78.47&82.70&58.70&84.54&78.37&11.56&28.88&26.42&\textbf{10.08}&34.52&15.34&21.13          \\

FedBN\cite{li2020fedbn}   &\underline{86.58}&79.83&78.16&77.93&\textbf{74.31}&\underline{87.01}&80.64&\textbf{8.65}&94.83&\underline{17.64}&74.84&26.24&\underline{10.09}&38.72 \\

FedLC\cite{wang2022personalizing} &84.07&80.96&77.09&82.39&73.10&79.17&79.46&10.40&26.20&24.85&10.42&31.28&24.79&21.32  \\
FedALA\cite{zhang2023fedala} &85.52&\underline{84.84}&81.21&\underline{83.78}&71.32&\textbf{87.95}&82.44&\underline{9.36}&24.98&18.22&15.06&32.02&\textbf{9.41}&18.18  \\
FedICRA\cite{lin2023unifying}  &\textbf{87.68}&83.70&\underline{84.03}&82.08&71.58&85.88&\underline{82.49}&9.59&\underline{14.94}&18.09&14.57&\underline{21.13}&15.09&\underline{15.57}\\

FedLPPA (Ours) &85.68&\textbf{86.88}&\textbf{87.91}&\textbf{86.41}&\underline{73.39}&86.60&\textbf{84.48}&12.09&\textbf{12.60}&\textbf{12.47}&14.13&\textbf{20.68}&10.72&\textbf{13.78}\\
\hline

LT (Weak)\cellcolor{gray!20}     &86.57\cellcolor{gray!20}&41.20\cellcolor{gray!20}&68.14\cellcolor{gray!20}&22.82\cellcolor{gray!20}&60.83\cellcolor{gray!20}&79.30\cellcolor{gray!20}&59.81\cellcolor{gray!20}&9.71\cellcolor{gray!20}&189.16\cellcolor{gray!20}&28.57\cellcolor{gray!20}&183.98\cellcolor{gray!20}&37.81\cellcolor{gray!20}&16.76\cellcolor{gray!20}&77.67\cellcolor{gray!20}   \\

CT (Weak)\cellcolor{gray!20}    &82.21\cellcolor{gray!20}&79.87\cellcolor{gray!20}&66.61\cellcolor{gray!20}&70.72\cellcolor{gray!20}&71.17\cellcolor{gray!20}&82.48\cellcolor{gray!20}&75.51\cellcolor{gray!20}&11.33\cellcolor{gray!20}&27.97\cellcolor{gray!20}&30.80\cellcolor{gray!20}&31.92\cellcolor{gray!20}&27.38\cellcolor{gray!20}&12.78\cellcolor{gray!20}&23.70\cellcolor{gray!20}     \\
LT (Full)\cellcolor{gray!20}     &90.92\cellcolor{gray!20}&88.36\cellcolor{gray!20}&89.93\cellcolor{gray!20}&73.96\cellcolor{gray!20}&86.10\cellcolor{gray!20}&91.48\cellcolor{gray!20}&86.79\cellcolor{gray!20}&7.99\cellcolor{gray!20}&11.35\cellcolor{gray!20}&9.42\cellcolor{gray!20}&40.52\cellcolor{gray!20}&20.79\cellcolor{gray!20}&7.92\cellcolor{gray!20}&16.33\cellcolor{gray!20}     \\
CT (Full)\cellcolor{gray!20}     &89.04\cellcolor{gray!20}&89.19\cellcolor{gray!20}&89.21\cellcolor{gray!20}&82.55\cellcolor{gray!20}&87.48\cellcolor{gray!20}&90.42\cellcolor{gray!20}&87.98\cellcolor{gray!20}&7.74\cellcolor{gray!20}&11.43\cellcolor{gray!20}&11.87\cellcolor{gray!20}&10.53\cellcolor{gray!20}&13.03\cellcolor{gray!20}&7.28\cellcolor{gray!20}&10.31\cellcolor{gray!20}    \\
\specialrule{0.12em}{0pt}{0pt}

\end{tabular}
}
\vspace{-0.4cm}
\end{table}

\begin{table}[b]
    \vspace{-0.4cm}
        \caption{{Performance comparisons of typical FL methods using our dual-decoder WSS paradigm on FAZ and polyp segmentation. The values in parentheses indicate the differences in performance metrics compared to using the WSS paradigm with only sparse label for supervision; {\color{cyan}cyan} denotes improvement, while {\color{orange}orange} denotes degradation.}}
        \label{new-table1}
        \centering
        \setlength{\tabcolsep}{0.9mm}
        \vspace{-0.15cm}
        \resizebox{\columnwidth}{!}{
        \begin{tabular}{c|cccc}
        \specialrule{0.12em}{0pt}{0pt}
        \multirow{2}{*}{Methods} & \multicolumn{2}{c}{FAZ} & \multicolumn{2}{c}{Polyp} \\
                             &           DSC$\uparrow$        & HD95$\downarrow$       & DSC$\uparrow$         & HD95$\downarrow$        \\
                             \hline
                             FedAvg \cite{mcmahan2017communication} & 88.85 ({\color{cyan}+4.80}) & 10.44 ({\color{cyan}-3.74}) & 77.62 ({\color{cyan}+16.35}) & 67.22 ({\color{cyan}-8.43}) \\
                            FedProx \cite{li2020federated} & \underline{90.11} ({\color{cyan}+2.57}) & \underline{8.17} ({\color{cyan}-0.17}) & \underline{77.66} ({\color{cyan}+14.36}) & 60.45 ({\color{cyan}-5.50}) \\
                            FedBN \cite{li2020fedbn} & 85.66 ({\color{cyan}+0.77}) & 9.51 ({\color{cyan}-1.76}) & 67.30 ({\color{cyan}+14.14})  & \underline{51.87} ({\color{cyan}-53.44})\\
                            FedRep \cite{collins2021exploiting} & 86.89 ({\color{cyan}+1.41}) & 12.51 ({\color{cyan}-1.09}) & 56.58 ({\color{cyan}+1.48})& 80.91 ({\color{cyan}-3.67})\\
                            FedALA \cite{zhang2023fedala} & 89.58 ({\color{cyan}+3.02})& 8.64 ({\color{orange}+0.74}) & 75.66 ({\color{cyan}+4.74}) & 67.50 ({\color{cyan}-4.98})\\
                             FedLPPA (Ours) & \textbf{93.51} & \textbf{5.30} & \textbf{80.99} & \textbf{37.60} 
                             \\
                            \specialrule{0.12em}{0pt}{0pt}
                                         
        \end{tabular}
        }
\end{table}

\begin{figure}[b]
    \vspace{-0.6cm}
    \centerline{\includegraphics[width=0.84\columnwidth]{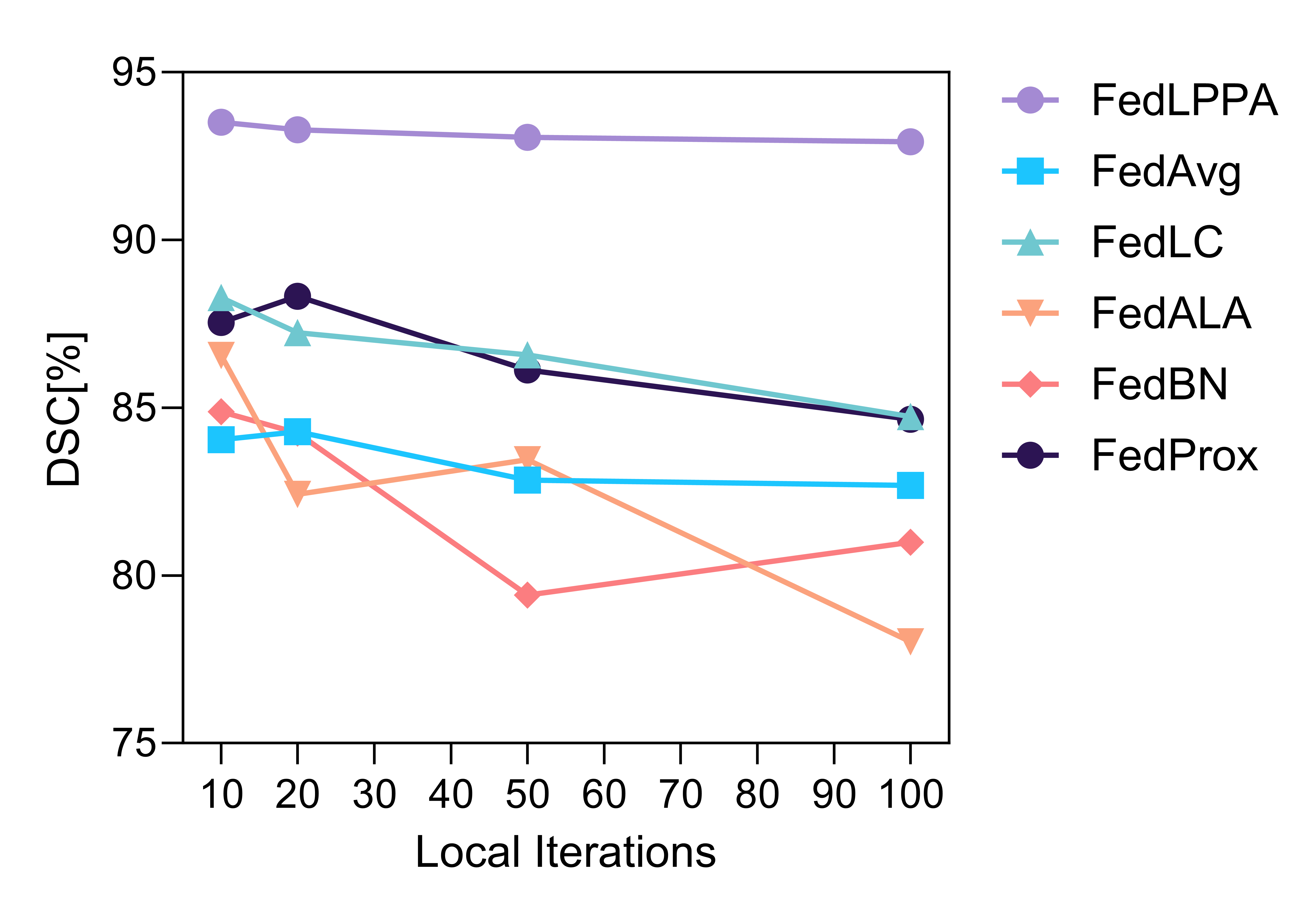}}
    \vspace{-0.3cm}
    \caption{{Performance of representative FL methods under various local iterations for the FAZ segmentation task.}}
    \label{fig6-new}
\end{figure}

\begin{figure*}[t]
    \vspace{-0.1cm}
    \centering
    \includegraphics[width=0.9\textwidth]{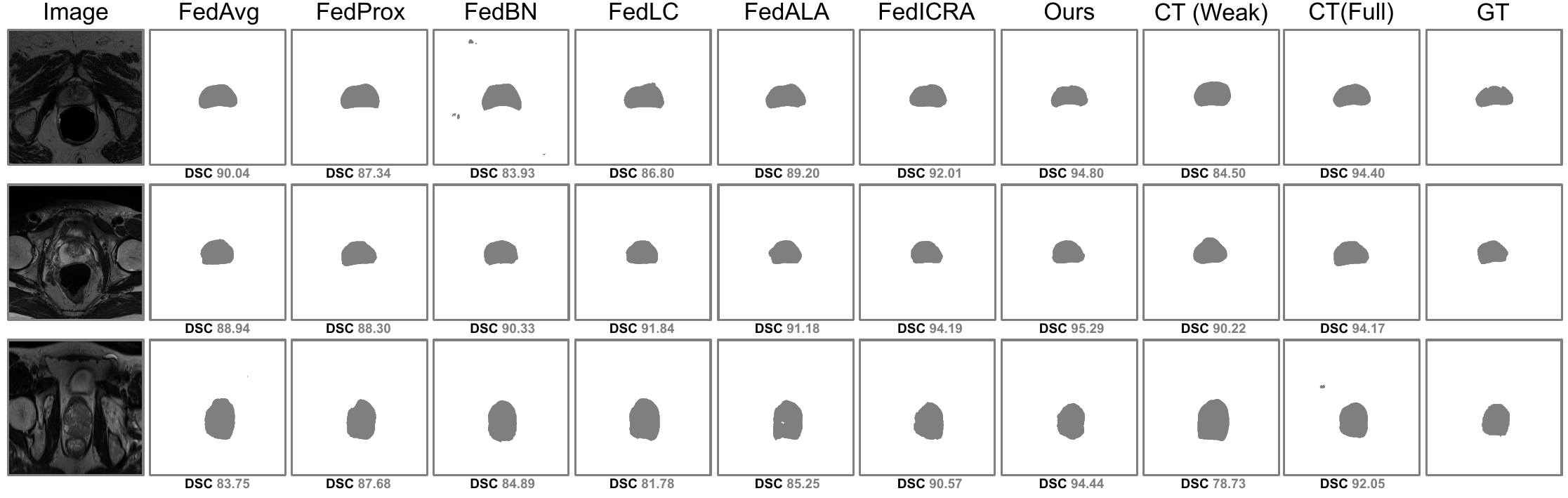}
    \vspace{-0.25cm}
    \caption{{Visualization results from FedLPPA and other SOTA methods on prostate segmentation (\textbf{CT} denotes centralized training, with \textbf{Weak} and \textbf{Full} respectively indicating the use of sparse annotations and full masks for model training).}}
    \label{fig4-new}
\end{figure*}

{
Considering that the communication efficiency is a crucial aspect of FL methods, we analyze this aspect by selecting several representative FL methods for comparative experiments with local iterations of 10, 20, 50, and 100 on the FAZ segmentation task—where a higher number of local iterations signifies lower communication frequency. The results, as displayed in Fig. \ref{fig6-new}, show that most FL methods exhibit performance degradation as communication frequency decreases. Notably, our FedLPPA exhibits the minimal performance decline and consistently outperforms all other methods at various communication frequencies.

To more fairly compare performance and underscore the effectiveness of our proposed framework, we implement the WSS paradigm used in FedLPPA for generating pseudo-proposal for several representative FL (including gFL and pFL) methods (employing the same set of loss functions, with the auxiliary decoder initialized differently from the main decoder as well as updated according to their respective aggregation methods or personalization strategies). We conduct experiments and performance comparisons on the FAZ and Polyp tasks, with results shown in Table \ref{new-table1}. The results confirm that applying our WSS design to typical FL methods enhances performance to a certain extent, yet FedLPPA still maintains a substantial performance advantage.}

\vspace{-0.1cm}
\subsection{Ablation Studies and Analyses}
We conduct detailed ablation analyses to investigate the effectiveness of each key component in FedLPPA. Subsequently, a series of intra-component ablations and analyses are performed to study their impact on the final performance.
\begin{table}[b]
\vspace{-0.35cm}
    \caption{
        Ablation study on the three key elements in FedLPPA. The columns for ODOC report the average metric values of OD and OC. {TDF: Tri-prompt Dual-attention Fusion, PD: Prompt similarity Dual-decoder, LA: Learnable Aggregation.}}
    \label{table5}
    \centering
    \setlength{\tabcolsep}{0.9mm}
    \vspace{-0.15cm}
    \resizebox{\columnwidth}{!}{
    \begin{tabular}{ccc|cccccc}
    \specialrule{0.12em}{0pt}{0pt}
    \multirow{2}{*}{TDF} & \multirow{2}{*}{PD} & \multirow{2}{*}{LA} & \multicolumn{2}{c}{ODOC} & \multicolumn{2}{c}{FAZ} & \multicolumn{2}{c}{Polyp} \\
                         &                     &                     & DSC$\uparrow$        & HD95$\downarrow$        & DSC$\uparrow$        & HD95$\downarrow$       & DSC$\uparrow$         & HD95$\downarrow$        \\
                         \hline
                         -  &     -                &     -                &    86.14        &   18.49          &    84.05        &    14.18        &    61.27         &   75.65          \\
                         \checkmark   &     -                &     -                &    86.31        &    15.83         &    90.61        & 7.14           &    72.40         &      75.01       \\
                         \checkmark &    \checkmark                 &    -                 &     89.68       &    11.07         &  92.49          &   6.97         &  77.58           &  49.25           \\
                         \checkmark &     -                &   \checkmark                  &     89.86       &   14.22          &   89.55         &   10.02         &  78.85           & 45.47 \\
                         \checkmark&\checkmark&\checkmark&90.76 &9.42 & 93.51 &5.30 & 80.99 &37.60 \\
                        \specialrule{0.12em}{0pt}{0pt}
                                     
    \end{tabular}
    }
    \end{table}
\vspace{-0.4cm}
\subsubsection{Ablation Studies of FedLPPA}
To delve deeper into FedLPPA, we perform ablations on the TDF module and the PDLA mechanism, where PDLA is divided into two parts: Prompt similarity Dual-decoder (PD, conducted on the server side) and Learnable Aggregation (LA, carried out on the client side). Table \ref{table5} presents the quantitative results of the main ablation studies across the {ODOC, FAZ, and Polyp} tasks. The first row represents the baseline, indicating a reduction to the classic FedAvg by removing the three components. The prompts in the TDF module are a prerequisite for acquiring the parameters of the auxiliary decoder. Hence, by initially integrating this module, a certain degree of performance improvement can be observed across all three tasks. In the primary ablation experiments, PD employs the {\textbf{Prompt Similarity Aggregated}} strategy mentioned in Sec. \ref{subsec:PDLA}. The inclusion of PD further significantly enhances performance on the previously established basis. The combination of the LA mechanism with TDF also yields commendable performance, but due to the absence of a dual-decoder to obtain reliable pseudo-proposals, it is not as effective as the TDF and PD combination on the ODOC and FAZ tasks (notably, a higher HD95 indicates a larger number of outlier predictions). With all components integrated, the optimal performance is achieved, underscoring the indispensability of each component.

\subsubsection{Ablation Studies of TDF}
In the ablation of the TDF module, we first retain the dual-attention fusion block and meticulously ablate the three types of prompts used in FedLPPA to explore the impact of each specific prompt. As DDP is used for measuring the data distribution of each client and forms the basis of the PD mechanism, it remains a constant element in this ablation experiment. From Table \ref{table6}, it is observable that the addition of either ASP or UKP results in a certain improvement in performance, and the incorporation of both to form the triple prompt composition of FedLPPA achieves the optimal performance outcomes.

\begin{table}[!t]
    \vspace{-0.25cm}
    \caption{
        Ablation study on the three prompts utilized in FedLPPA. DDP: data distribution
        prompt, ASP: annotation sparsity prompt, UKP: universal knowledge prompt.}
    \vspace{-0.15cm}
    \centering
    \setlength{\tabcolsep}{0.85mm}
    \label{table6}
    \resizebox{\columnwidth}{!}{
    \begin{tabular}{ccc|cccccc}
    \specialrule{0.12em}{0pt}{0pt}
    \multirow{2}{*}{DDP} & \multirow{2}{*}{ASP} & \multirow{2}{*}{UKP} & \multicolumn{2}{c}{ODOC} & \multicolumn{2}{c}{FAZ} & \multicolumn{2}{c}{Polyp} \\
                         &                     &                     & DSC$\uparrow$        & HD95$\downarrow$        & DSC$\uparrow$        & HD95$\downarrow$       & DSC$\uparrow$         & HD95$\downarrow$        \\
                         \hline
                         \checkmark   &     -                &     -                &    88.66       &     16.38      &    92.68        &    6.20        &    78.33         &  44.68           \\
                         \checkmark &    \checkmark                 &    -                 &      88.91       &     14.15   &  93.18          &    5.39        &  78.50           &  42.56           \\
                         \checkmark &     -                &   \checkmark                  &     90.02       &     9.77        &    92.91        &    5.61        &  79.26           & 38.50 \\
                         \checkmark&\checkmark&\checkmark&90.76 &9.42 & 93.51 & 5.30 & 80.99 &37.60 \\
                        \specialrule{0.12em}{0pt}{0pt}
                                     
    \end{tabular}}
    \end{table}

\begin{figure}[tbp]
        \vspace{-0.15cm}
        \centerline{\includegraphics[width=\columnwidth]{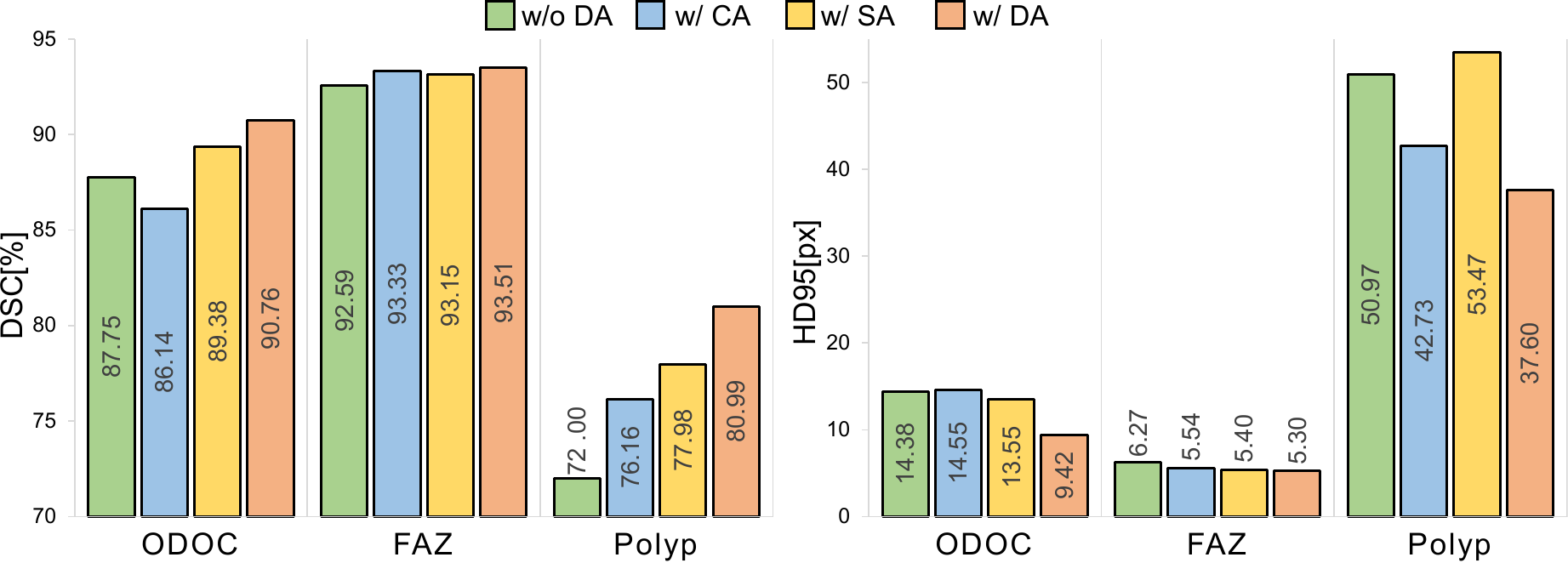}}
        \vspace{-0.1cm}
        \caption{Performance of FedLPPA utilizing various feature fusion and enhancement blocks. \textbf{CA}, \textbf{SA} and \textbf{DA} respectively denote channel attention, spatial attention and dual attention fusion blocks.}
        \label{fig5}
        \vspace{-0.5cm}
    \end{figure}

\begin{figure}[htbp]
    \vspace{-0.1cm}
    \centerline{\includegraphics[width=\columnwidth]{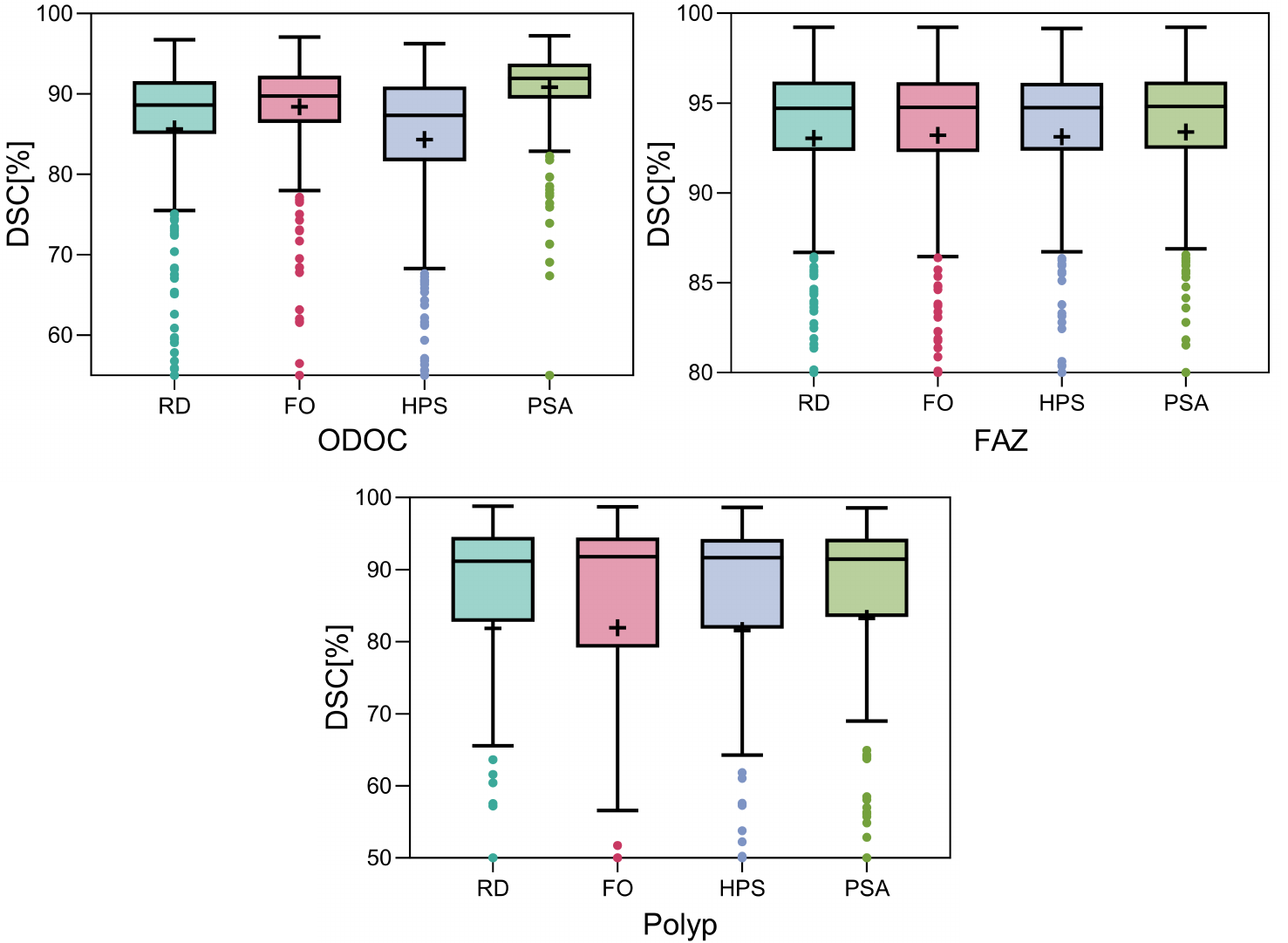}}
    \vspace{-0.2cm}
    \caption{{Performance of FedLPPA utilizing various server-side auxiliary decoder parameter selection strategies. \textbf{RD}, \textbf{FO}, \textbf{HPS} and \textbf{PSA} respectively denote Random, Fixed Order, Highest Prompt Similarity and Prompt Similarity Aggregated strategies.}}
    \label{fig6}
    \vspace{-0.4cm}
\end{figure}

Given that the original triple prompts consist of only a limited number of bits, simply concatenating the prompts with sample features and processing them through several convolutional layers often proves inadequate for effective feature fusion and enhancement. Consequently, the effectiveness of the dual attention fusion block is further explored. Fig. \ref{fig5} demonstrates that variants without any specific type of feature fusion nor enhancement achieve lower results in most metrics across the three {(ODOC, FAZ and Polyp)} tasks. Implementing either channel attention (CA) fusion or spatial attention (SA) fusion generally leads to performance improvements in most cases. However, exceptions are noted, such as that on the ODOC task where only using CA results in a performance decline. 
This could be attributed to the high data heterogeneity across different centers, where each local model is already capable of making personalized adjustments to various features, and the relatively coarse granularity of CA alone leads to a slight performance reduction. Yet, the application of dual attention fusion for feature enhancement consistently yields the best results in both types of metrics across the three tasks. This underscores the necessity of finer-grained modulation, fusion and enhancement of the triple prompts with sample features. 
\subsubsection{Comparisons of Server-side Auxiliary Decoder Parameter Selection Strategies}
In Sec. \ref{subsec:PDLA}, we present four strategies for obtaining the auxiliary decoder parameters on the server side, which will be distributed to the clients. These strategies include {\textbf{Random} (RD), \textbf{Fixed Order} (FO), \textbf{Highest Prompt Similarity} (HPS), and \textbf{Prompt Similarity Aggregated} (PSA, used as the default)}. In this subsection, we first explore the performance of these strategies and present the results in Fig. \ref{fig6}. 
The figure shows that the performance differences among the four strategies are not very significant across the three tasks. However, the PSA strategy consistently achieves the best performance in all scenarios. This is because the other three strategies only utilize information from a single separate client, while PSA aggregates and utilizes information from multiple similar sites based on prompt-calculated similarity. It also filters out information from dissimilar sites.

\begin{figure}[tbp]
    
    \hspace{+1.15cm} PCA \hspace{+1.75cm} t-SNE \hspace{+0.85cm} Correlation Matrix
    
    \leftline{
    \hspace{+0.06cm}\includegraphics[width=0.3045\columnwidth]{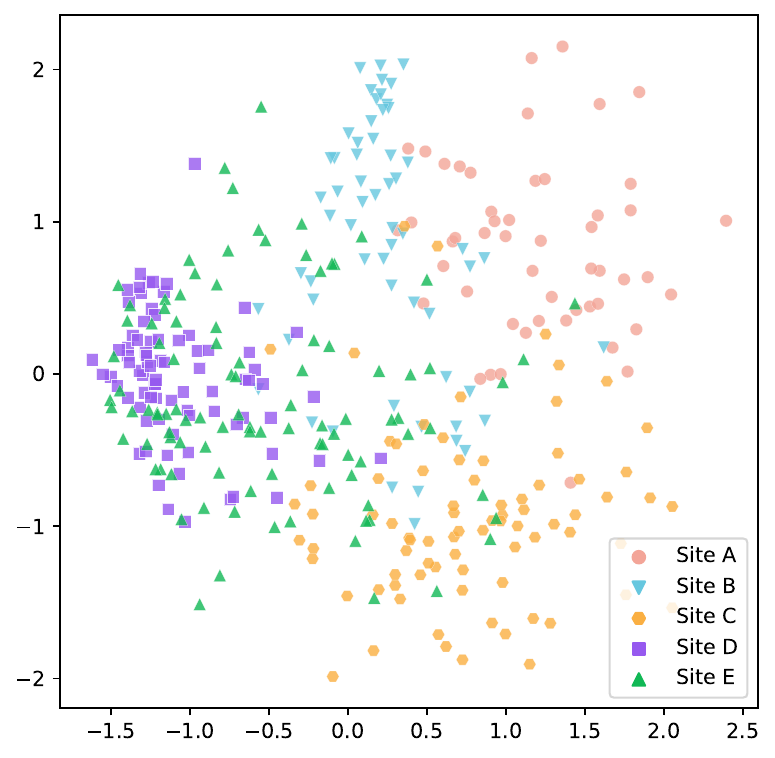}\hspace{-0.1cm}
    \includegraphics[width=0.31\columnwidth]{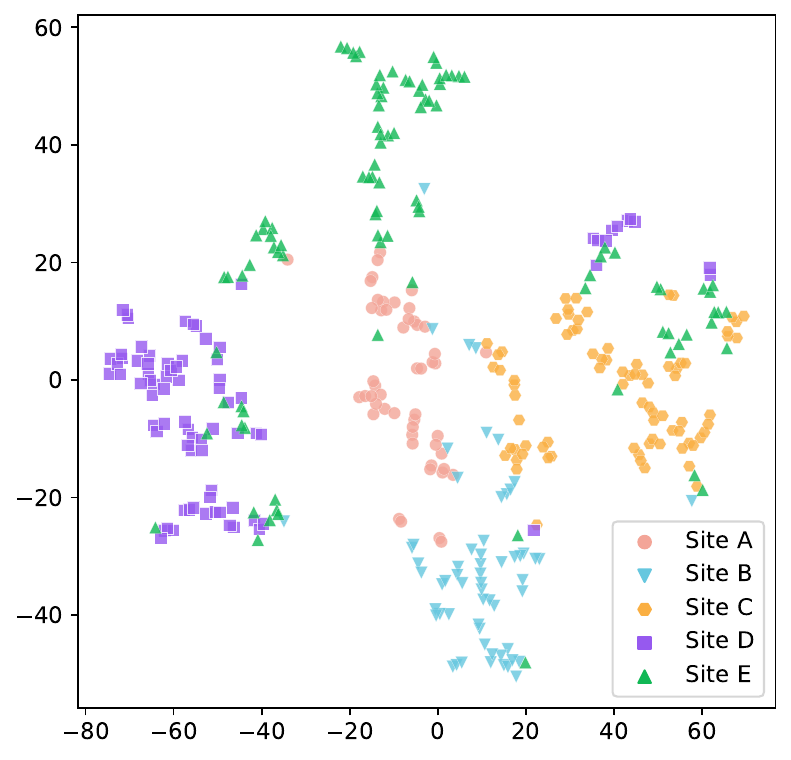}\hspace{-0.1cm}
    \includegraphics[width=0.375\columnwidth]{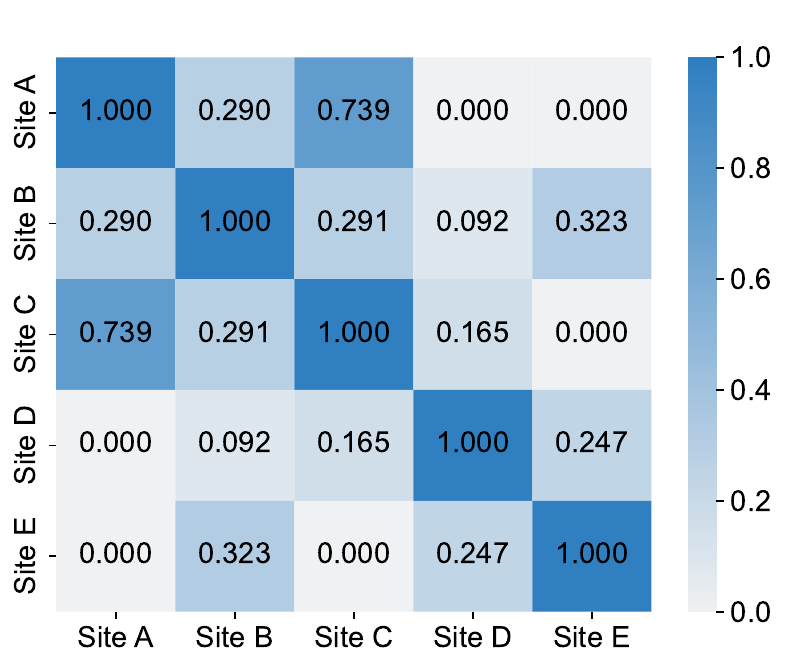}
    }
    
    \centering{(A)}
    \vspace{-0.1cm}

    \leftline{
    \includegraphics[width=0.31\columnwidth]{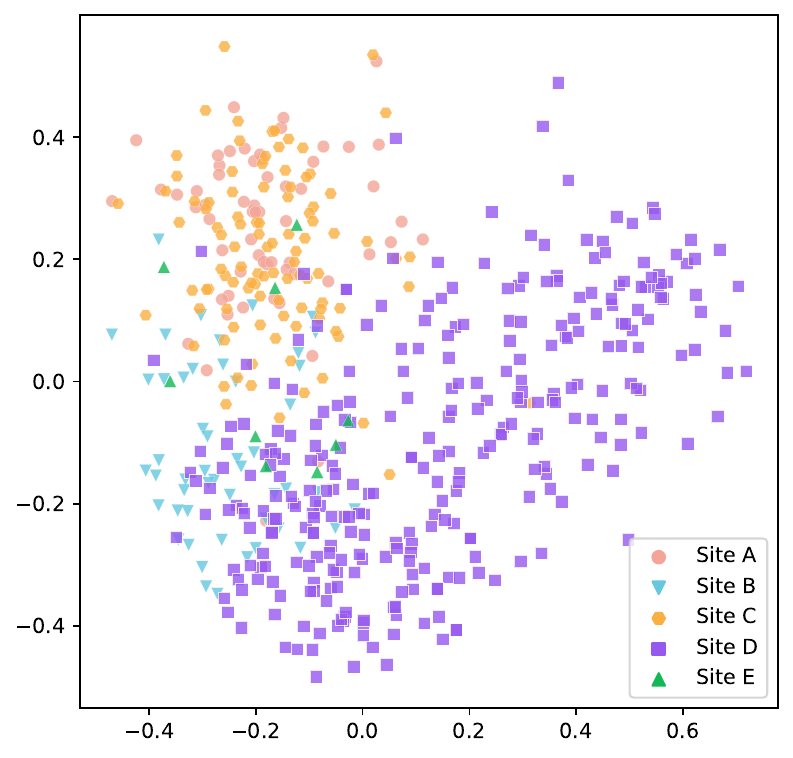}\hspace{-0.1cm}
    \includegraphics[width=0.31\columnwidth]{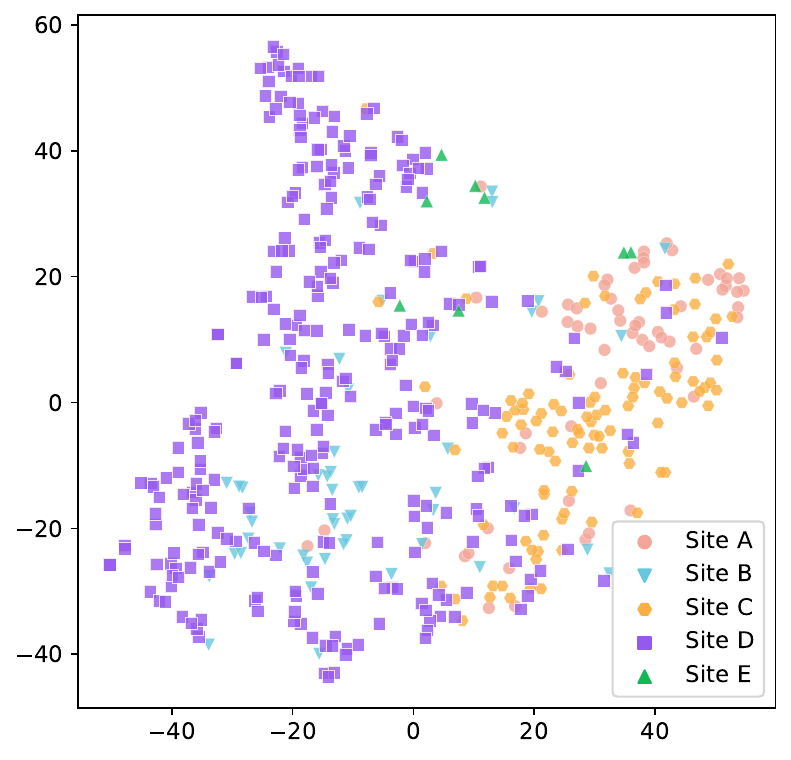}\hspace{-0.1cm}
    \includegraphics[width=0.375\columnwidth]{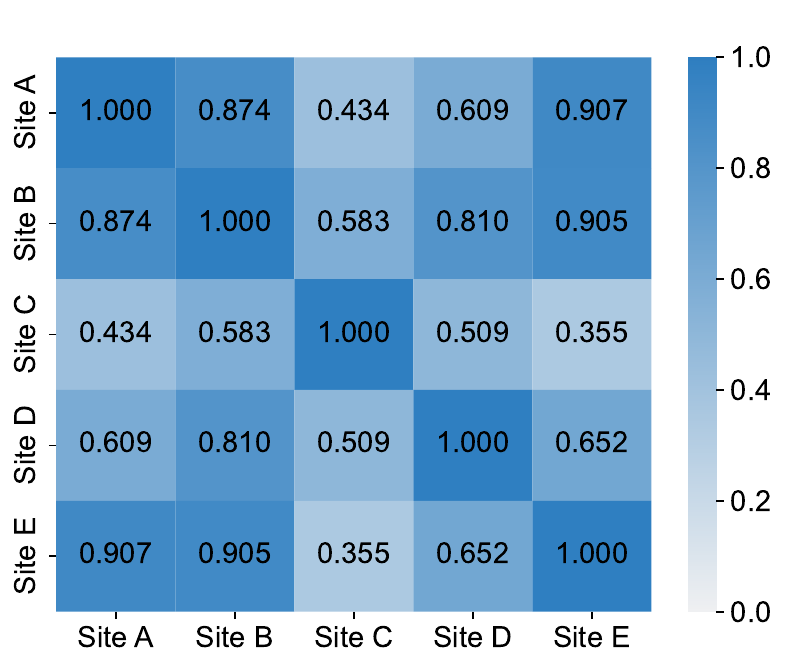}}
    
    \centering{(B)}
    \vspace{-0.1cm}
    
    \leftline{
    \includegraphics[width=0.31\columnwidth]{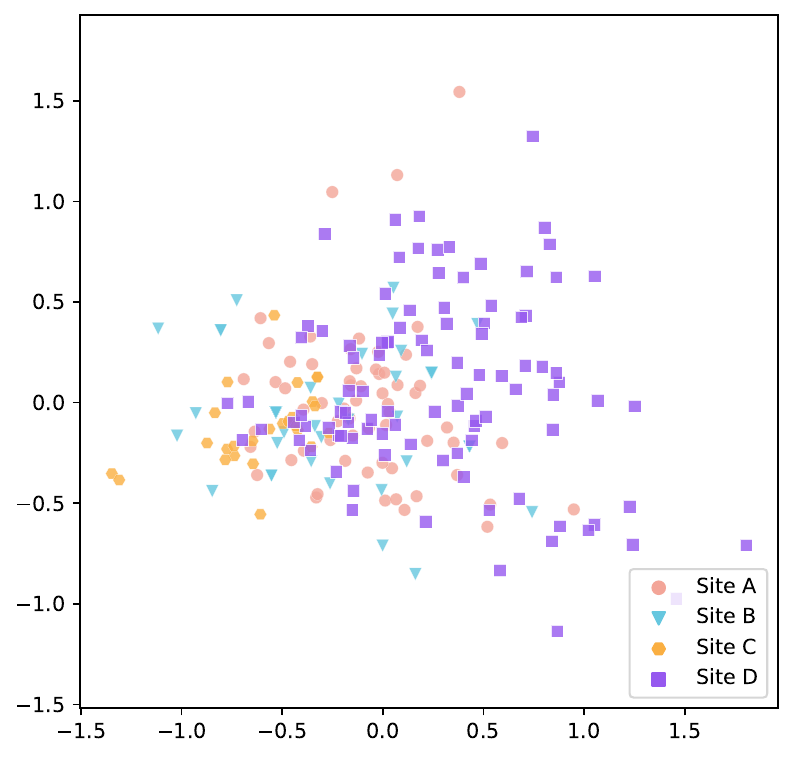}\hspace{-0.1cm}
    \includegraphics[width=0.31\columnwidth]{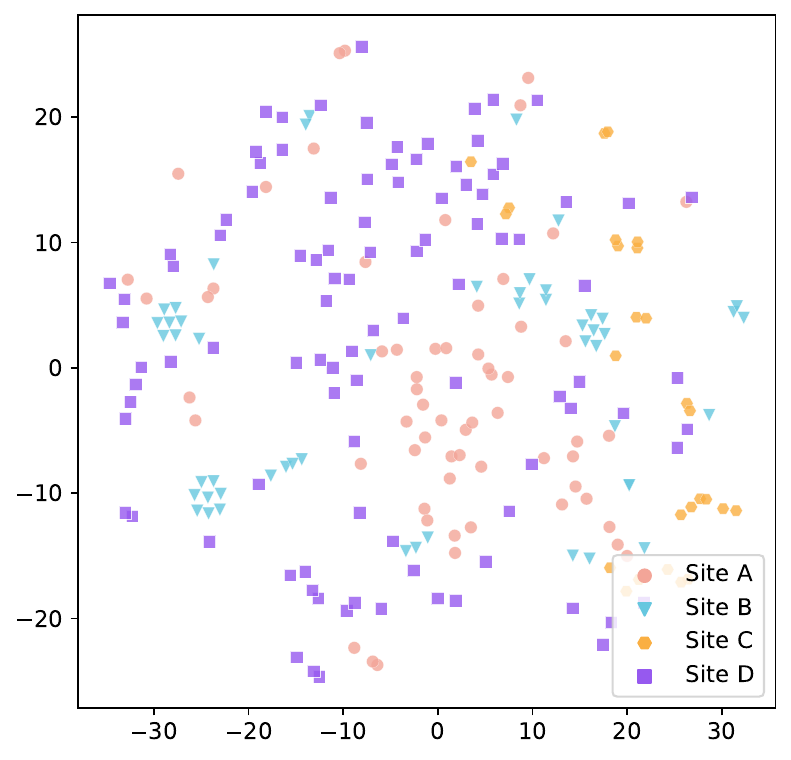}\hspace{-0.1cm}
    \includegraphics[width=0.375\columnwidth]{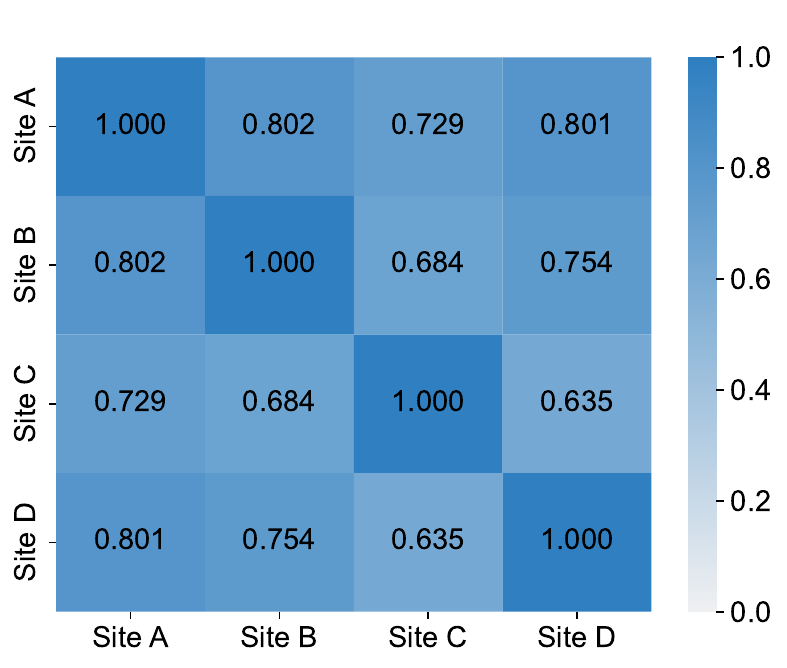}
    }
    
    \centering{(C)}

  \vspace{-0.1cm}
  \caption{{The PCA, t-SNE visualization results, and the correlation matrices for three ((A) ODOC, (B) FAZ, and (C) Polyp) segmentation tasks.}}
  \vspace{-0.45cm}
  \label{fig:tsne&cor} 
  \end{figure}

\subsubsection{Affinity Matrix Analysis}
In Fig. \ref{fig:tsne&cor}, we perform {principal component analysis (PCA) and} t-SNE visualization (using the encoded features from the segmentation models trained in a centralized learning manner) for three {(ODOC, FAZ, and Polyp)} segmentation tasks, as well as the {inter-client} similarity/affinity in each scenario (derived from our FedLPPA framework), calculated based on UKP and DDP. From the corresponding {PCA, }t-SNE and affinity matrices, strong correlation between data heterogeneity and affinity degree is observed. For example, in the ODOC task, where data heterogeneity among clients is high, the corresponding matrix shows lower similarity values and has zeros (indicating dissimilar clients). In contrast, in the FAZ and Polyp scenarios, where data heterogeneity is lower, the values in the affinity matrices are higher, especially in the Polyp task. The Polyp data has the least heterogeneity and all values in the corresponding affinity matrix are greater than 0.5. Furthermore, although not strictly corresponding, the relative distribution/relationship of data features in {PCA, }t-SNE and the similarities obtained from FedLPPA demonstrate a high degree of correlation, validating the effectiveness of using prompts to calculate similarities among multi-center data. Additionally, in conjunction with the results in Sec. \ref{sota exp}, it clearly illustrates that FedLPPA effectively achieves personalization under various data heterogeneity scenarios, demonstrating its broad applicability, generalizability and robust performance.

\subsubsection{{Model and Communication Efficiency}}

{
In Table \ref{table9}, we present the local model parameter size, communication overhead required per communication round, training time per federated round, and testing efficiency for various FL methods. It is important to note that these metrics for different FL methods are provided under identical hardware and communication conditions. Although FedLPPA shows a slight decrease in efficiency compared to other FL methods, its trade-off in efficiency is not significant relative to the performance improvement it offers.
}

\begin{table}[h]
    \vspace{-0.1cm}
    \caption{
        {Comparison of model and communication efficiency. U: Upload, D: Download, \textit{N}: Number of Clients.}}
    \label{table9}
    \centering
    \setlength{\tabcolsep}{0.5mm}
    \vspace{-0.15cm}
    \resizebox{\columnwidth}{!}{
    \begin{tabular}{l|cccc}
    \specialrule{0.12em}{0pt}{0pt} 

    Methods  & \makecell{Model \\Parmas (M)} & \makecell{Communication \\Overhead (M)} & \makecell{Training \\Time (min/round)} & \makecell{Test Efficiency \\(samples/s)} \\
    \hline
    FedAvg                                                                                       & 1.8135             & [1.8135 (U) + 1.8135 (D)]$\times N$ & 0.200  & 43          \\

    FT                                                                                    & 1.8135             & [1.8135 (U) + 1.8135 (D)]$\times N$  & 0.200  & 43          \\

    FedProx                                                                                      & 1.8135             & [1.8135 (U) + 1.8135 (D)]$\times N$  & 0.724  & 43          \\

    FedBN                                                                                        & 1.8135             & [1.8105 (U) + 1.8105 (D)]$\times N$  & 0.204  & 43          \\

    FedAp                                                                                        & 1.8135             & [1.8105 (U) + 1.8105 (D)]$\times N$  & 0.654  & 43          \\

    FedRep                                                                                       & 1.8135             & [1.8132 (U) + 1.8132 (D)]$\times N$ & 0.202  & 43          \\

    MetaFed                                                                                      & 1.8135             & [1.8105 (U) + 1.8105 (D)]$\times N$  & 0.722  & 43          \\

    FedLC                                                                                        & 2.1109             & [2.1099 (U) + 2.1099 (D)]$\times N$  & 0.634  & 36          \\

    FedALA                                                                                       & 1.8135             & [1.8135 (U) + 1.8135 (D)]$\times N$  & 0.456  & 43          \\

    FedICRA                                                                                      & 2.3372             & [2.3372 (U) + 2.3372 (D)]$\times N$  & 0.582  & 30          \\

    FedLPPA                                                                                      & 2.8978             & [2.2564 (U) + 2.8978 (D)]$\times N$  & 0.710  & 29   \\
    \specialrule{0.12em}{0pt}{0pt}

    \end{tabular}}
    \end{table}
    \vspace{-0.2cm}

\section{Conclusion}

This paper presents a novel pFL framework, FedLPPA, for tackling a practical but challenging federated heterogeneous WSS problem. A TDF module and a PDLA mechanism are introduced to enable in-context learning and fine-grained personalization. Extensive experiments are conducted on {four} medical image segmentation tasks, with superiority of FedLPPA being successfully established. Additionally, we have shown that under heterogeneous WSS conditions, FedLPPA can achieve segmentation performance that is comparable to that of labor-intensive fully-supervised centralized training.




\bibliography{tmi}
\bibliographystyle{IEEEtran}

\end{document}